\documentclass[runningheads]{llncs}

\usepackage{eccv}

\usepackage{eccvabbrv}

\usepackage{graphicx}
\usepackage{booktabs}
\usepackage{wrapfig}
\usepackage{multirow}
\usepackage{color, xcolor, colortbl}

\usepackage[accsupp]{axessibility}  % Improves PDF readability for those with disabilities.

\usepackage{hyperref}

\usepackage{orcidlink}

\usepackage{hhline} % used for the top-most horizontal line! in order to fix the top left corner cell empty
\usepackage{highlight}
\begin{document}

\title{Benchmarking Spurious Bias in Few-Shot Image Classifiers} 
% BSB
%TODO REVIEW: If the paper title is too long for the running head, you can set

\author{Guangtao Zheng\inst{1}\orcidlink{0000-0002-1287-4931} \and
Wenqian Ye\inst{1}\orcidlink{0000-0002-6069-5153} \and
Aidong Zhang\inst{1}\orcidlink{0000-0001-9723-3246}}

\authorrunning{G.~Zheng et al.}

\institute{$^{1}$University of Virginia, Charlottesville VA 22904, USA\\
\email{\{gz5hp,wenqian,aidong\}@virginia.edu}}

\maketitle

\begin{abstract}
Few-shot image classifiers are designed to recognize and classify new data with minimal supervision and limited data but often show reliance on spurious correlations between classes and spurious attributes, known as spurious bias.
Spurious correlations commonly hold in certain samples and few-shot classifiers can suffer from spurious bias induced from them. There is an absence of an automatic benchmarking system to assess the robustness of few-shot classifiers against spurious bias.  
In this paper, we propose a systematic and rigorous benchmark framework, termed FewSTAB, to fairly demonstrate and quantify varied degrees of robustness of few-shot classifiers to spurious bias. FewSTAB creates few-shot evaluation tasks with biased attributes so that using them for predictions can demonstrate poor performance. To construct these tasks, we propose attribute-based sample selection strategies based on a pre-trained vision-language model, eliminating the need for manual dataset curation. This allows FewSTAB to automatically benchmark spurious bias using any existing test data.  FewSTAB offers evaluation results in a new dimension along with a new design guideline for building robust classifiers. Moreover, it can benchmark spurious bias in varied degrees and enable designs for varied degrees of robustness. Its effectiveness is demonstrated through experiments on ten few-shot learning methods across three datasets. We hope our framework can inspire new designs of robust few-shot classifiers. Our code is available at \url{https://github.com/gtzheng/FewSTAB}.

\end{abstract}
%FewSTAS ASBench    
\section{Introduction}

% \begin{wrapfigure}[19]{R}{0.5\linewidth}
% \centering
%     \includegraphics[width=0.95\linewidth]{figures/shortcut_illustration.pdf}%
%   \caption{Exploiting the spurious correlation between the class \texttt{bird} and the spurious attribute \texttt{tree branch} to predict \texttt{bird} leads to an incorrect prediction on the test image showing birds on a grass field. For clarity, we only show the case for one class.}%
%     \label{fig:shortcut-illustration}%
% \end{wrapfigure}

\begin{figure}[t]
\centering
    \includegraphics[width=\linewidth]{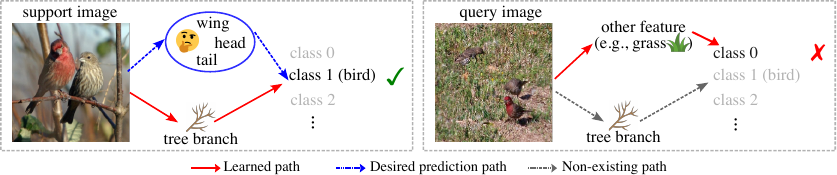}%
  \caption{Exploiting the spurious correlation between the class \texttt{bird} and the spurious attribute \texttt{tree branch} to predict \texttt{bird} leads to an incorrect prediction on the test image showing birds on a grass field. For clarity, we only show the case for one class.}%
    \label{fig:shortcut-illustration}%
\end{figure}

Few-shot classification \cite{Snell2017Prototypical,bertinetto2018metalearning,lee2019meta,ye2020fewshot,zhang2020deepemdv2} (FSC) has attracted great attention recently due to its promise for recognizing novel classes efficiently with limited data. 
Few-shot classifiers can transfer the knowledge learned from base classes to recognize novel classes with a few labeled samples. However, they face potential risks when deployed in the real world, such as data distribution shifts \cite{motiian2017few,yue2021multi} and adversarial examples \cite{goldblum2020adversarially,dong2022improving}. 
A subtle yet critical risk factor is the spurious correlations \cite{geirhosimagenet,stock2018convnets,baker2018deep,beery2018recognition,sagawadistributionally,ghosal2024vision,xue2023eliminating} between classes and spurious attributes --- attributes of inputs non-essential to the classes. In the traditional learning setting \cite{arjovsky2019invariant,sagawadistributionally,ahmed2020systematic}, deep learning models tend to rely on spurious correlations as their prediction shortcuts or exhibit
\textit{spurious bias}, such as predicting
classes using the associated backgrounds \cite{xiao2021noise} or image textures \cite{geirhos2018imagenettrained}, 
leading to significant performance drops when the associated backgrounds or textures change to different ones. 
In the low-data regime, spurious bias becomes more evident.
For example, in Fig. \ref{fig:shortcut-illustration}, the correlation between the class \texttt{bird} and the attribute \texttt{tree branch} in the support (training) image may form a shortcut path from \texttt{tree branch} to 
predicting the image as \texttt{bird} and hinder the learning of the desired one that uses class-related attributes, such as \texttt{head}, \texttt{tail}, and \texttt{wing}. The shortcut will fail to generalize in the query (test) image where no \texttt{tree branch} can be found. In general, few-shot image classifiers are susceptible to spurious bias.

However, there lacks a dedicated benchmarking framework that evaluates the robustness of few-shot classifiers to spurious bias. The standard benchmarking procedure in FSC trains a few-shot classifier on base classes from a training set with ample samples and evaluates the classifier on FSC test tasks constructed from a test set with novel classes. The problem with this procedure is the lack of explicit control over the spurious correlations in the constructed FSC tasks. Each FSC test task contains randomly sampled support and query samples. Thus, spurious correlations in the majority of the test set samples can be demonstrated in these tasks, providing unfair advantages for few-shot classifiers with high reliance on the spurious correlations. 
%sTherefore, the existing benchmarking method cannot fairly compare different few-shot classifiers on their robustness to spurious bias. 

 In this paper, we propose a systematic and rigorous benchmark framework, termed Few-Shot Tasks with Attribute Biases (FewSTAB), to fairly compare the robustness of various few-shot classifiers to spurious bias. Our framework explicitly controls spurious correlations in the support and query samples when constructing an FSC test task to reveal the robustness pitfalls caused by spurious bias. 
 To achieve this, we propose attribute-based sample selection strategies that select support and query samples with biased attributes. These attributes together with their associated classes formulate spurious correlations such that if the support samples induce spurious bias in a few-shot classifier, i.e., the classifier learns the spurious correlations in the support samples as its prediction shortcuts, then the query samples can effectively degrade the classifier's performance, exposing its non-robustness to spurious bias.

Our framework exploits the spurious attributes in test data for formulating spurious correlations in FSC test tasks. Some existing datasets \cite{sagawadistributionally,he2021towards,liu2015deep} provide spurious attribute annotations. However, they only have a few classes and cannot provide enough classes for training and testing. Many benchmark datasets for FSC do not have annotations on spurious attributes, and obtaining these annotations typically involves labor-intensive human-guided labeling \cite{nushi2018towards,zhang2018manifold}.  To address this, we further propose to use a pre-trained vision-language model (VLM) to automatically identify distinct attributes in images in the high-level text format. Our attribute-based sampling methods can use the identified attributes to simulate various spurious correlations.  Thus, we can reuse any existing FSC datasets for benchmarking few-shot classifiers' robustness to spurious bias, eliminating the need for the manual curation of new datasets.

The main contributions of our work are summarized as follows:
\begin{itemize}
    \item We propose a systematic and rigorous benchmark framework, termed Few-Shot Tasks with Attribute Biases (FewSTAB), that specifically targets spurious bias in few-shot classifiers, demonstrates their varied degrees of robustness to spurious bias, and benchmarks spurious bias in varied degrees. 
    \item We propose novel attribute-based sample selection strategies using a pre-trained VLM for constructing few-shot evaluation tasks,  allowing us to reuse any existing few-shot benchmark datasets without manually curating new ones for the evaluation.
    \item FewSTAB provides a new dimension of evaluation on the robustness to spurious bias along with a new design guideline for building robust few-shot classifiers. We demonstrate the effectiveness of FewSTAB by applying it to models trained on three benchmark datasets with ten FSC methods.
\end{itemize}
%without creating new datasets?

% To create test tasks that are realistic in simulating the few-shot learning and inference scenarios in real world and  can faithfully reflect the few-shot learning capability of a model, we propose a novel concept-based sampling method 

\section{Related Work}

\textbf{Few-shot classification}. Few-shot classification \cite{Vinyals2016Matching,Snell2017Prototypical,chen2018a,wang2020generalizing,tian2020rethinking,Triantafillou2020Meta-Dataset} has received vast attention recently. Few-shot classifiers can be trained with meta-learning or transfer learning on base classes to learn the knowledge that can be transferred to recognize novel classes with a few labeled samples.  The transfer learning approaches \cite{chen2018a,tian2020rethinking} first learn a good embedding model and then fine-tune the model on samples from novel classes. The meta-learning approaches can be further divided into optimization-based and metric-based methods. The optimization-based methods \cite{finn2017model,Li2017MetaSGDLT,Lee2018GradientBasedMW,Wang2018Low,Flennerhag2020MetaLearning} aim to learn a good initialized model such that the model can adapt to novel classes efficiently with a few gradient update steps on a few labeled samples. The metric-based methods \cite{Vinyals2016Matching,Sung2018Learning,rusu2018metalearning,lee2019meta,Oreshkin2018TADAM,bertinetto2018metalearning,ye2020fewshot,Zhang_2020_CVPR} aim to learn a generalizable representation space with a well-defined metric, such as Euclidean distance \cite{Snell2017Prototypical}, to learn novel classes with a few labeled samples.  Recently, large vision-language models \cite{radford2021learning,zhu2023prompt,khattak2023maple} are used for few-shot classification. However, they have completely different training and inference pipelines from the models that we consider in this paper.

\noindent\textbf{Robustness in few-shot classification}. There are several notions of robustness for few-shot classifiers. The common one requires a few-shot classifier to perform well on the in-distribution samples of novel classes in randomly sampled FSC test tasks. The robustness to adversarial perturbations further requires a few-shot classifier to perform well on samples with imperceptible perturbations \cite{goldblum2020adversarially,dong2022improving}. Moreover, the cross-domain generalization \cite{Tseng2020CrossDomain,Triantafillou2020Meta-Dataset,qin2023bi} aims to test how robust a few-shot classifier is on samples from novel classes with domain shifts, which are typically reflected by the changes in both image styles and classes. In contrast, we focus on a new notion of robustness: the robustness to spurious bias.
There is a lack of rigorous evaluation methods on the topic. We provide
a new evaluation method that specifically targets spurious bias and can systematically demonstrate few-shot classifiers' varied degrees of vulnerability to spurious bias, which has not been addressed in the existing literature.

\noindent\textbf{Benchmarks for spurious bias}. There are some existing datasets \cite{sagawadistributionally,he2021towards,liu2015deep} that are designed to benchmark spurious bias in image classifiers. However, these datasets are only applicable to the traditional learning setting \cite{arjovsky2019invariant,sagawadistributionally,ahmed2020systematic} since the classes in them are not sufficient for the training and testing of few-shot classifiers. Existing benchmarks in few-shot classification are not tailored for benchmarking spurious bias in few-shot classifiers. A recent work \cite{zhang2024metacoco} creates a large-scale few-shot classification benchmark dataset with spurious-correlation shifts. In contrast, we propose a benchmark framework that can reuse existing few-shot classification datasets and provide a new dimension of evaluation.
%on the robustness of few-shot classifiers to spurious bias.

\noindent\textbf{Discovering spurious attributes}. A spurious attribute is non-essential to a class and only exists in some samples. Early works on discovering spurious attributes \cite{nushi2018towards,zhang2018manifold} require a predefined list of spurious attributes and expensive human-guided labeling of visual attributes. Recent works \cite{singla2021understanding,wong2021leveraging,singlasalient,neuhaus2022spurious} greatly reduce the need for manual annotations by using the neurons of robust models to detect visual attributes. However, they still need humans to annotate the detected visual attributes.  We automate this process by using a pre-trained VLM to obtain distinct attributes as words. Instead of discovering spurious correlations, we simulate them via attribute-based sampling for benchmarking.

\section{Preliminary}\label{sec:preliminary}
%In this section, we introduce the basics and the problem settings of few-shot classification. For clarity, we list major symbols used in the paper alongside their meanings in Table \ref{tab:symbol-table}.

\noindent\textbf{Few-shot classification tasks.} 
A typical FSC task $\mathcal{T}$ has a support set $\mathcal{S}$ for training and a query set $\mathcal{Q}$ for testing. In this task, there are $C$ classes ($c=1, \ldots,C$) with $N_\mathcal{S}$ (a small number) training samples and $N_\mathcal{Q}$ test samples per class in $\mathcal{S}$ and $\mathcal{Q}$, respectively. The task is called a $C$-way $N_\mathcal{S}$-shot task. 
%Classes and the samples within these classes may vary across different tasks.

\begin{table}[t]
\centering
\caption{Meanings of major symbols used in the paper.}
\resizebox{0.55\columnwidth}{!}{%
\begin{tabular}{ll}
\toprule
\multicolumn{1}{c}{Symbol} & \multicolumn{1}{c}{Meaning} \\ \midrule
$\mathcal{T}$ & An FSC task \\
$\mathcal{S}$ & Support (training) set in $\mathcal{T}$ \\
$\mathcal{Q}$ & Query (test) set in $\mathcal{T}$ \\
$c$ & A class in $\mathcal{T}$\\
$C$ & Number of  classes per task \\
$N_\mathcal{S}$ & Number of samples (shots) per class in $\mathcal{S}$ \\
$N_\mathcal{Q}$ & Number of samples per class in $\mathcal{Q}$ \\
$\mathcal{O}$ & A few-shot adaptation algorithm \\
$\psi$ & An attribute detector \\
$\Omega$ & An automatic word selector \\
$\mathcal{D}_{train}$ & The base training set \\
$\mathcal{D}_{val}$ & The validation set for selecting  a few-shot classifier\\
$\mathcal{D}_{test}$ & The test set for evaluating a few-shot classifier \\
$\mathcal{C}_{train}$ & Classes in $\mathcal{D}_{train}$ \\
$\mathcal{C}_{test}$ & Classes in $\mathcal{D}_{test}$ \\ 
$\mathcal{D}_{c}$ & A set of all samples belonging to class $c$ \\
$a$ & A text-format attribute\\
$\mathcal{A}$ & A set of text-format attributes \\ \bottomrule
\end{tabular}%
}
\label{tab:symbol-table}
\end{table}

% \begin{wraptable}[18]{r}{7cm}
% \centering
% \resizebox{0.55\columnwidth}{!}{%
% \begin{tabular}{ll}
% \toprule
% \multicolumn{1}{c}{Symbol} & \multicolumn{1}{c}{Meaning} \\ \midrule
% $\mathcal{T}$ & An FSC task \\
% $\mathcal{S}$ & Support (training) set in $\mathcal{T}$ \\
% $\mathcal{Q}$ & Query (test) set in $\mathcal{T}$ \\
% $c$ & A class in $\mathcal{T}$\\
% $C$ & Number of  classes per task \\
% $N_\mathcal{S}$ & Number of samples (shots) per class in $\mathcal{S}$ \\
% $N_\mathcal{Q}$ & Number of samples per class in $\mathcal{Q}$ \\
% $\mathcal{O}$ & A few-shot adaptation algorithm \\
% $\psi$ & A feature detector \\
% $\Omega$ & An automatic word selector \\
% $\mathcal{D}_{train}$ & The base training set \\
% $\mathcal{D}_{val}$ & The validation set for selecting  a few-shot classifier\\
% $\mathcal{D}_{test}$ & The test set for evaluating a few-shot classifier \\
% $\mathcal{C}_{train}$ & Classes in $\mathcal{D}_{train}$ \\
% $\mathcal{C}_{test}$ & Classes in $\mathcal{D}_{test}$ \\ 
% $\mathcal{D}_{c}$ & A set of all samples belonging to class $c$ \\
% $a$ & A text-format attribute\\
% $\mathcal{A}$ & A set of text-format attributes \\ \bottomrule
% \end{tabular}%
% }
% \caption{Meanings of major symbols used in the paper.}
% % \vspace{-5mm}
% \label{tab:symbol-table}
% \end{wraptable}

\noindent\textbf{Few-shot classifiers.}  A few-shot classifier $f_\theta$ with parameters $\theta$ aims to classify the samples in $\mathcal{Q}$ after learning from $\mathcal{S}$ with a learning algorithm $\mathcal{O}$ in a few-shot task $\mathcal{T}$.
Here, $\mathcal{O}$ could be any learning algorithms, such as the optimization method \cite{finn2017model} or a prototype-based classifier learning method \cite{Snell2017Prototypical,chen2020new}. To acquire a good few-shot learning capability,  $f_\theta$ is typically meta-trained or pre-trained \cite{li2021LibFewShot} on a base training set $\mathcal{D}_{train}=\{(x_n,y_n)|y_n\in\mathcal{C}_{train},n=1,\ldots,N_{train}\}$ with $N_{train}$ sample($x$)-label($y$) pairs, where $\mathcal{C}_{train}$ is a set of base classes.

\noindent\textbf{Performance metrics.} The performance of a few-shot classifier is typically measured by its average classification accuracy over $N_\mathcal{T}$ $C$-way $N_\mathcal{S}$-shot tasks randomly sampled from $\mathcal{D}_{test}=\{(x_n,y_n)|y_n\in\mathcal{C}_{test},n=1,\ldots,N_{test}\}$ where $N_{test}$ sample-label pairs from novel classes $\mathcal{C}_{test}$ do not appear in $\mathcal{D}_{train}$, \ie, $\mathcal{C}_{train}\cap\mathcal{C}_{test}=\emptyset$.
We denote this metric as \textit{standard accuracy} $\text{Acc}(f_\theta)$, \ie,
\begin{align}\label{eq:standard-accuracy}%\vspace{-2mm}
\text{Acc}(f_\theta)=\frac{1}{N_\mathcal{T}}\sum_{t=1}^{N_\mathcal{T}}\sum_{c=1}^CM_c(\mathcal{T}_t;f_\theta, \mathcal{O}),
\end{align}
where $M_c(\mathcal{T}_t;f_\theta, \mathcal{O})$ denotes the classification accuracy of $f_\theta$ on the query samples from the class $c$ in $\mathcal{T}_t$ after $f_\theta$ is trained on $\mathcal{S}$ with $\mathcal{O}$. 
The metric Acc$(f_\theta)$ in Eq. \eqref{eq:standard-accuracy} only shows the average learning capability of $f_\theta$ over  $C$ randomly selected novel classes.  To better characterize the robustness of $f_\theta$  to spurious bias, we define the \textit{class-wise worst classification accuracy} over tasks as 
\begin{align}\label{eq:worst-accuracy}%\vspace{-2mm}
\text{wAcc}(f_\theta)=\frac{1}{N_\mathcal{T}}\sum_{t=1}^{N_\mathcal{T}}\min_{c=1,\ldots,C}M_c(\mathcal{T}_t;f_\theta, \mathcal{O}).
\end{align}
A larger $\text{wAcc}(f_\theta)$ indicates that  $f_{\theta}$ is  more robust to spurious bias.

\noindent\textbf{Spurious correlations}. A spurious correlation is the association between a class and an attribute of inputs that is \textit{non-essential} to the class, and it \textit{only} holds in some samples. We formally define it as follows.

\begin{definition}
   Let $\mathcal{D}_c$ denote a set of sample-label pairs having the label $c$, and let $\psi:\mathcal{X}\rightarrow \mathcal{B}_\mathcal{A}$ be an attribute detector, where $\mathcal{X}$ is the set of all possible inputs, $\mathcal{B}_\mathcal{A}$ denotes all possible subsets of $\mathcal{A}$, and $\mathcal{A}$ is the set of all possible attributes.  The class $c$ and an attribute $a\in\mathcal{A}$ form a spurious correlation, denoted as $\langle c,a\rangle$, if and only if the following conditions hold:
   \begin{enumerate}
       \item There exists $(x, c) \in \mathcal{D}_c$ that satisfies $a \in \psi(x)$, and
       \item There exists $(x', c) \in \mathcal{D}_c$ that satisfies $a \notin \psi(x')$.
   \end{enumerate}
  We define $a$ as the \textbf{spurious attribute} in $\langle c,a\rangle$.
   \label{def:spurious-correlation}
\end{definition} 

% For example, we can find two spurious correlations in Fig. \ref{fig:shortcut-illustration}: $\langle$\texttt{bird}, \texttt{tree branch}$\rangle$ and $\langle$\texttt{bird}, \texttt{grass}$\rangle$, with \texttt{tree branch} and \texttt{grass} being the spurious attributes.
\cref{def:spurious-correlation} specifies that all the spurious correlations are based on $\mathcal{D}_c$.
In the remainder of the paper, we define $\mathcal{D}_c=\{(x,c)|\forall(x,c)\in\mathcal{D}_{test}\}$ with $c\in\mathcal{C}_{test}$ as we focus on \textit{evaluating} the robustness to spurious bias.

We list major symbols in the paper alongside their meanings in \cref{tab:symbol-table}.
% \begin{definition}[\textbf{Spurious Correlation}]
%    Given a class $c$ and a feature $a$, we say that $c$ and $a$ forms a spurious correlation denoted as $\langle c,a\rangle$ if $\exists x\in\mathcal{D}_c$, $a\in\psi(x)$, and $\exists x'\in\mathcal{D}_c$, $a\notin\psi(x')$, where $\psi$ is a general feature detector, and $\mathcal{D}_c$ represents the set of all samples belonging to class $c$. We call $a$ spurious attribute in a spurious correlation $\langle c,a\rangle$. \label{def:spurious-correlation}
% \end{definition}

\begin{figure}[t]
    \centering
	\includegraphics[width=\linewidth]{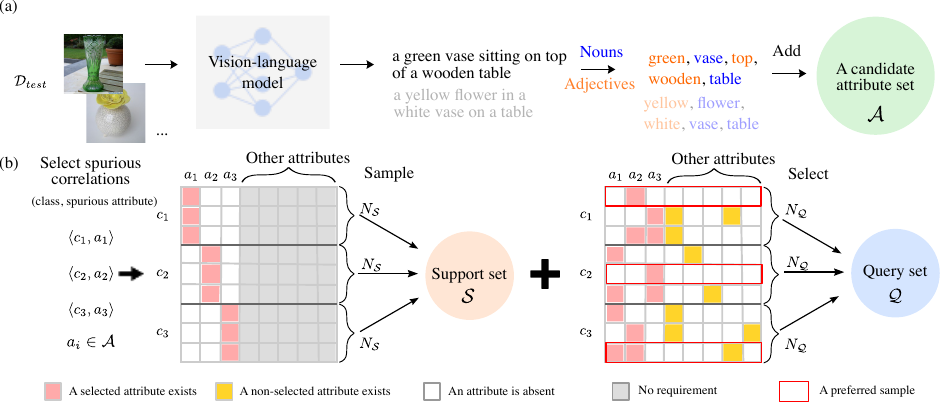}
	\caption{FewSTAB overview. (a) Extract  distinct attributes using a pre-trained VLM. (b) Generate an FSC task for the evaluation of spurious bias in few-shot classifiers.
	}
	\label{fig:method-overview}
 % \vspace{-2mm}
\end{figure}

\section{Methodology}
% We first propose two attribute-based sample selection methods to reveal specific spurious bias (Section \ref{sec:spurious-attack}). Then, we propose a systematic and rigorous benchmark framework, termed \textbf{FewSTAB},  that performs automated attribute detection with a pre-trained VLM (Section \ref{sec:spurious-feature-detection}) and constructs FSC tasks to benchmark spurious bias in few-shot classifiers (Section \ref{sec:spurious-test-task}).

\subsection{Attribute-Based Sample Selection}\label{sec:spurious-attack}
We first propose two attribute-based sample selection methods to reveal spurious bias in a few-shot classifier.
Consider a training set $\mathcal{S}$ in a few-shot test task $\mathcal{T}$, which has $C$ classes with each class $c\in\mathcal{C}_{test}$ associating with a unique spurious attribute $a\in\mathcal{A}$. We aim to discover samples that can exhibit a classifier's spurious bias on $\langle c,a\rangle$ induced from $\mathcal{S}$.
%i.e., the classifier will generalize poorly on the samples if using $\langle c,a\rangle$ for predictions.
Motivated by existing findings \cite{sagawadistributionally,xue2023eliminating,yang2023mitigating} that classifiers with high reliance on $\langle c,a\rangle$ tend to perform poorly on samples without it, we propose an attribute-based sample selection strategy
below.%, termed \textit{intra-class attribute-based sample selection}, which is defined as follows.

\noindent \textbf{Intra-class attribute-based sample selection}. Given $\mathcal{D}_c$ and the training set $\mathcal{S}$ having the spurious correlation $\langle c,a\rangle$, we generate a set $\mathcal{I}_{\langle c,a\rangle}$ of sample-label pairs which have class $c$ but do not contain attribute $a$, i.e.,
\begin{align}\label{eq:intra-class-attack}
 \mathcal{I}_{\langle c,a\rangle} = \{(x, c)|\forall (x, c) \in \mathcal{D}_c, a \notin \psi(x)\}.
 \end{align}

% \begin{align}\label{eq:intra-class-attack}
%  AS_{intra}(c, \mathcal{D}_{test}) = \{(x, c)|\forall(x,c)\in\mathcal{D}_{test}, a \notin \psi(x)\}.
% \end{align}

% \end{definition}
% \begin{definition}[\textbf{attribute bias}]\label{def:spurious-attack}
% A attribute bias on the spurious correlation $\langle c,a\rangle$ is a subset of samples $\mathcal{D}_c^{att}$ from the set of samples with class label $c$, $\mathcal{D}_c$: 
% \begin{equation}
% \begin{aligned}
%     &\mathcal{D}_c^{att} = \{(x, c)|\forall (x, c) \in \mathcal{D}_c, a \notin \psi(x)\} \\
%     &\cap \{(x, c) | \forall (x, c) \in \mathcal{D}_c, \exists a' \in \psi(x), a' \not\equiv c\}
% \end{aligned}
% \end{equation}
% \noindent Here, $\not\equiv$ indicates no direct semantic alignment between $a'$ and $c$. For instance, as illustrated in Fig. \ref{fig:shortcut-illustration}, the concept of \texttt{tree branch} is not semantically aligned with \texttt{bird}."
% \end{definition}

The above proposed method demonstrates a few-shot classifier's robustness to individual spurious 
correlation $\langle c,a\rangle$ and does not consider a multi-class classification setting where spurious attributes from some other class $c'$ exist in samples of the class $c$. In this case, these attributes may mislead the classifier to predict those samples as the class $c'$ and severely degrade the performance on the class $c$. For example, consider using the spurious correlations $\langle$\texttt{vase}, \texttt{blue}$\rangle$ and $\langle$\texttt{bowl}, \texttt{green}$\rangle$ for predicting \texttt{vase} and \texttt{bowl}, respectively. An image showing a vase in green is more effective in revealing the robustness to  $\langle$\texttt{vase}, \texttt{blue}$\rangle$ as it is more likely to be misclassified as \texttt{bowl} than other images. %showing a vase in other colors.
Motivated by this, we propose the \textit{inter-class attribute-based sample selection} below.

\noindent \textbf{Inter-class attribute-based sample selection.} Given  $\mathcal{D}_c$ and the training set $\mathcal{S}$ having the spurious correlations $\langle c,a\rangle$ and $\langle c',a'\rangle$,  where $c'\neq c$ and $a'\neq a$, we generate a set  $\mathcal{I}_{\langle c,a\rangle}^{\langle c',a'\rangle}$ of sample-label pairs which have class $c$, do not contain attribute $a$, but contain attribute $a'$ from another class $c'$:
\begin{align}\label{eq:cls-attack-set}
 \mathcal{I}_{\langle c,a\rangle}^{\langle c',a'\rangle} = \mathcal{I}_{\langle c,a\rangle}\cap \{(x, c)|\forall (x, c) \in \mathcal{D}_c, a'_{\langle c' \rangle} \in \psi(x)\},
 \end{align}%
 where $a'_{\langle c' \rangle}$ denotes $a'$ in $\langle c',a'\rangle$, and $\mathcal{I}_{\langle c,a\rangle}$ is defined in \cref{eq:intra-class-attack}.% is from the intra-class attribute-based sample selection.

Considering that there are $C$ classes in the training set $\mathcal{S}$ with each class associating with a unique spurious attribute $a$, to effectively demonstrate the reliance on the spurious correlation $\langle c,a\rangle$ with the inter-class attribute-based sample selection, we consider all the spurious correlations in $\mathcal{S}$.  Specifically, we apply the above selection strategy to all the $C-1$ spurious correlations in $\mathcal{S}$ other than $\langle c,a\rangle$ and obtain $\mathcal{I}_{\langle c,a\rangle}^{C}$ as the union of the $C-1$ sets as follows:
 \begin{equation}\label{eq:multi-cls-attack-set}
     \mathcal{I}_{\langle c,a\rangle}^{C} = \bigcup_{\langle c',a'\rangle\in\mathcal{C}_{\backslash c}}\mathcal{I}_{\langle c,a\rangle}^{\langle c',a'\rangle},
 \end{equation}
 where $\mathcal{C}_{\backslash c}$ denotes all the spurious correlations in $\mathcal{S}$ other than $\langle c,a\rangle$.

The inter-class attribute-based sample selection is built upon the intra-class attribute-based sample selection. In the remainder of the paper, we use the inter-class method as our default sample selection strategy, which is more effective empirically (\cref{sec:ablation}). In certain cases, however, where there are not enough desired samples during task construction, we resort to the intra-class sample selection strategy (\cref{sec:experiment-setup}, Implementation details).

 %We select samples from $\mathcal{D}_{test}$ in an existing few-shot classification dataset, which does not require generating new data that may lie out of the original data distributions in $\mathcal{D}_{test}$.

 In the following, we introduce \textbf{FewSTAB}, a benchmark framework that uses the proposed selection strategies to construct FSC tasks containing samples with biased attributes for benchmarking spurious bias in few-shot classifiers.
 
 % Given the training samples $\mathcal{S}$ in a $C$-way $N_\mathcal{S}$-shot task $\mathcal{T}$,  our sample selection generate query samples in $\mathcal{Q}$ as $\cup_{c=1}^C\mathcal{Q}_c$ with $\mathcal{Q}_c$ containing $N_\mathcal{Q}$ samples randomly selected from the above attacks.  

\subsection{FewSTAB (Part 1): Text-Based Attribute Detection}\label{sec:spurious-feature-detection}
Our attribute-based sample selection methods require knowing the attributes in images, which typically involves labor-intensive human labeling. To make our method scalable and applicable to few-shot classifiers trained on different datasets, we adopt a pre-trained VLM to automatically identify distinct attributes in images in text format, which includes the following two steps.

\noindent\textbf{Step 1: Generating text descriptions.} We use a pre-trained VLM \cite{nlp_connect_2022,li2022blip} $\phi$ to automatically generate text descriptions for images in $\mathcal{D}_{test}$. The VLM is a model in the general domain and can produce text descriptions for various objects and patterns.  For example, for the current input image in the \texttt{vase} class in  \cref{fig:method-overview}(a),  besides the class object \texttt{vase}, the VLM also detects the vase's color \texttt{green}, and another object \texttt{table} with its material \texttt{wooden}.

\noindent\textbf{Step 2: Extracting informative words.} From the generated text descriptions, we extract nouns and adjectives as the detected attributes via  \textit{an automatic procedure} $\Omega$.  The two kinds of words are informative as a noun describes an object, and an adjective describes a property of an object. All the detected attributes form the candidate attribute set $\mathcal{A}$. We realize the attribute detector $\psi$ defined in \cref{def:spurious-correlation} as $\psi(x)=\Omega(\phi(x))$.

 \noindent\textbf{Remark 1:} A VLM in general can extract many distinct attributes from the images. On some images, the VLM may detect non-relevant attributes, such as detecting a duck from a bird image. A more capable VLM could warrant a better attribute detection accuracy and benefit individual measurements on few-shot classifiers. Although being a VLM-dependent benchmark framework, FewSTAB can produce consistent and robust relative measurements among all the compared FSC methods, regardless the choice of VLMs (\cref{sec:ablation}).

% The VLM $\psi$ may not generate the exact text descriptions from images. One example would be describing a lemon on a tree as a yellow bird. 
% However, this inconsistency in VLM does not significantly impact our framework's effectiveness.  Since the same set of extracted attributes will be used for the evaluation on all FSC methods, the comparison between these methods is valid and fair. While a more capable VLM could potentially warrant a better image-to-text performance, all the FSC methods are still evaluated and compared in the same setting with the same VLM. Thus the relative measurements of different FSC methods from our framework with different VLMs can converge. We demonstrate this argument with two different VLMs in Table \ref{tab:spearman-corr-vlm}.

\noindent\textbf{Remark 2:} The candidate set $\mathcal{A}$ constructed with all the extracted words may contain attributes that represent the classes in $\mathcal{D}_{test}$. However, during our attribute-based sample selection, these attributes will not be used since they always correlate with classes and therefore do not satisfy the definition of spurious attributes in \cref{def:spurious-correlation}.
 We provide details of $\phi$ and $\Omega$ in \cref{sec:experiment-setup}.

\subsection{FewSTAB (Part 2): FSC Task Construction}\label{sec:spurious-test-task}
Constructing a $C$-way $N_\mathcal{S}$-shot FSC task $\mathcal{T}$ for benchmarking spurious bias in few-shot classifiers involves constructing a support (training) set $\mathcal{S}$  and a query (test) $\mathcal{Q}$ with biased attributes.

 %To make the attack more effective, we also craft the spurious correlations in $\mathcal{S}$ to ``trick" the classifier into learning the spurious correlations as its prediction shortcuts. Therefore, generating a challenging FSC task includes the ``trick" and the ``attack" parts which correspond to generating $\mathcal{S}$ and $\mathcal{Q}$, respectively.

\noindent\textbf{Constructing the support set.} The support set contains the spurious correlations that we aim to demonstrate to a few-shot classifier. As a fair and rigorous benchmark system, FewSTAB makes no assumptions on the few-shot classifiers being tested and randomly samples $C$ classes from $\mathcal{C}_{test}$. For each sampled class, it randomly selects a spurious correlation $\langle c,a\rangle$ in $\mathcal{D}_{test}$ with $a\in\mathcal{A}$ . To effectively demonstrate the spurious correlation $\langle c,a\rangle$ to a few-shot classifier, we select samples of the class $c$ such that (1) they all have the spurious attribute $a$ and (2) do not have spurious attributes from the other $C-1$ spurious correlations. 
We construct $\mathcal{S}_c$ with $N_\mathcal{S}$ samples for the class $c$ that satisfy the above two conditions. Thus, the spurious attribute $a$ becomes predictive of the class $c$ in $\mathcal{S}_c$.  We take the union of all $C$ such sets to get  $\mathcal{S}=\cup_{c=1}^C\mathcal{S}_c$.
\cref{fig:method-overview}(b) demonstrates the case when $C=3$. Note that we have no requirements for other non-selected attributes in $\mathcal{A}$ to ensure that we have enough samples for $\mathcal{S}_c$.

\noindent\textbf{Constructing the query set.} To evaluate the robustness to the spurious correlations formulated in $\mathcal{S}$, we first construct a candidate set $\mathcal{I}_{\langle c,a\rangle}^{C}$ in Eq. \eqref{eq:multi-cls-attack-set} for each spurious correlation $\langle c,a\rangle$ in $\mathcal{S}$. Since we have no requirements on the non-selected attributes that are \textit{not} used to formulate spurious correlations in $\mathcal{S}$, a few-shot classifier may predict query samples via some of these attributes, e.g., the yellow blocks in \cref{fig:method-overview}(b), bypassing the test on the formulated spurious correlations in $\mathcal{S}$. To address this, we propose query sample selection below. \\
\textit{Query sample selection:}  We select query samples from $\mathcal{I}_{\langle c,a\rangle}^{C}$ that are \textit{least likely} to have non-selected spurious attributes, such as the ones enclosed with red boxes in \cref{fig:method-overview}(b). 
To achieve this, we first calculate the fraction of sample-label pairs in $\mathcal{I}_{\langle c,a\rangle}^{C}$ that have the attribute $a$ as
\begin{align}
    p_a=|\{x|a\in \psi(x),\forall (x,c)\in \mathcal{I}_{\langle c,a\rangle}^{C}\}|/|\mathcal{I}_{\langle c,a\rangle}^{C}|,
\end{align}
where $|\cdot|$ denotes the size of a set, $a\in\tilde{\mathcal{A}}$, and $\tilde{\mathcal{A}}$ contains all non-selected attributes. A larger $p_a$ indicates that the attribute $a$ occurs more frequently in data and is more likely to be used in formulating prediction shortcuts. We then calculate the likelihood score for each $(x,c)\in\mathcal{I}_{\langle c,a\rangle}^{C}$ as $s(x) = \sum_{a\in\psi(x),a\in\tilde{\mathcal{A}}} p_a$, i.e., the summation of all $p_a$ of non-selected attributes in $x$. The likelihood score will be zero if there are no non-selected attributes in $x$. A large $s(x)$ indicates that the image $x$ can be predicted via many non-selected attributes. Therefore, we select $N_\mathcal{Q}$ samples from $\mathcal{I}_{\langle c,a\rangle}^{C}$ that have the lowest likelihood scores to construct $\mathcal{Q}_c$. Then, we have $\mathcal{Q}=\cup_{c=1}^C\mathcal{Q}_c$, which contains samples for evaluating the robustness of a few-shot classifier to the spurious correlations in $\mathcal{S}$.

\noindent\textbf{Complexity analysis.} The text-based attribute detection only needs to use VLMs once to extract attributes for each test set of a
dataset. For the task construction, in a nutshell, we analyze the attributes of samples from each of the $C$ classes and do the sampling. Thus, the computational complexity is $O(N_\mathcal{T}CN_cN_{\mathcal{A}})$, where $N_c$ is the maximum number of samples per class in test data, $N_{\mathcal{A}}$ is the number of extracted attributes. We only need to run the process \textit{once} and use the generated tasks to benchmark various models.

\section{Experiments}
\subsection{Experimental Setup}\label{sec:experiment-setup}
\textbf{Datasets}. We used two general datasets of different scales, miniImageNet~\cite{Ravi2017OptimizationAA} and tieredImageNet~\cite{ren2018metalearning}, and one fine-grained dataset, CUB-200 \cite{WelinderEtal2010}. Each dataset consists of $\mathcal{D}_{train}$, $\mathcal{D}_{val}$, and $\mathcal{D}_{test}$ for training, validation, and test, respectively (see Appendix).
All images were resized to $84\times84$.

\noindent\textbf{FSC methods}. We trained FSC models with ten algorithms covering three major categories. For gradient-based meta-learning algorithms, we chose ANIL \cite{Raghu2020Rapid}, LEO \cite{rusu2018metalearning}, and BOIL \cite{oh2020boil}. For metric-based meta-learning algorithms, we chose ProtoNet \cite{Snell2017Prototypical}, DN4 \cite{li2019revisiting}, R2D2 \cite{bertinetto2018metalearning}, CAN \cite{hou2019cross}, and RENet \cite{kang2021relational}. For transfer learning algorithms, we chose Baseline++ \cite{chen2018a} and RFS \cite{tian2020rethinking}.
See Appendix for more details.
Any backbones can be used as the 
feature extractor. For fair comparisons between different methods, we used the ResNet-12 backbone adopted in \cite{Oreshkin2018TADAM}.

\noindent\textbf{Text-based attribute detection.} We used a pre-trained VLM named 
ViT-GPT2 \cite{nlp_connect_2022} to generate text descriptions for images in $\mathcal{D}_{test}$. 
After that, we used Spacy (\url{https://spacy.io/}) to extract nouns and adjectives from these descriptions automatically. 
We also used another pre-trained VLM, BLIP \cite{li2022blip}, to test whether FewSTAB can produce consistent results. 
The statistics of the detected attributes are shown in Table \ref{tab:feat-statistics}.

\begin{table}[t]
\centering
\caption{Statistics of detected attributes in $\mathcal{D}_{test}$ by two VLMs.}
\resizebox{0.9\linewidth}{!}{%
\begin{tabular}{ccccccc}
\hline
\multirow{2}{*}{VLM} & \multicolumn{3}{c}{Unique attributes}   & \multicolumn{3}{c}{Avg. attributes per class} \\ \cline{2-7} 
                     & miniImageNet & tieredImageNet & CUB-200 & miniImageNet   & tieredImageNet   & CUB-200   \\ \hline
ViT-GPT2             & 1111         & 2532           & 159     & 190.40         & 230.94           & 25.78     \\
BLIP                 & 2032         & 6710           & 247     & 254.40         & 310.40           & 29.74     \\ \hline
\end{tabular}%
}
\label{tab:feat-statistics}
\end{table}
% \begin{wraptable}{r}{0.43\linewidth}
% \centering
% %\vspace{-6mm}
% \caption{Statistics of detected attributes in $\mathcal{D}_{test}$ by two VLMs.}
% \resizebox{\linewidth}{!}{%
% \begin{tabular}{lccc}
% \toprule
% Dataset & VLM & \multicolumn{1}{c}{\begin{tabular}[c]{@{}c@{}}Unique \\ attributes\end{tabular}} & \begin{tabular}[c]{@{}c@{}}Avg. attributes \\ per class\end{tabular} \\ \midrule
% \multirow{2}{*}{miniImageNet} & ViT-GPT2 & 1111 & 190.40 \\
%  & BLIP & 2032 & 254.40 \\ \midrule
% \multirow{2}{*}{tieredImageNet} & ViT-GPT2 & 2532 & 230.94 \\
%  & BLIP & 6710 & 310.40 \\ \midrule
% \multirow{2}{*}{CUB-200} & ViT-GPT2 & 159 & 25.78 \\
%  & BLIP & 247 & 29.74 \\ \bottomrule
% \end{tabular}%
% }
% \label{tab:feat-statistics}
% \end{wraptable}

\noindent\textbf{Implementation details.} 
We trained FSC 
models with the implementation in \cite{li2021LibFewShot}. 
Each model was trained on $\mathcal{D}_{train}$ of a dataset with one of the ten FSC methods. For each meta-learning based method, we trained two models using randomly sampled 5-way 1-shot and 5-way 5-shot tasks, respectively. All the tasks have 15 samples per class in the query set. We saved the model that achieves the best validation accuracy on $\mathcal{D}_{val}$ for evaluation. 
For FewSTAB, if we do not have enough desired samples to construct a support set, we redo the construction from the beginning. If there are not enough desired samples to construct a query set, we first try to use the intra-class attribute-based sample selection; if the desired samples are still not enough, we redo the construction from the beginning.
We created 3000 tasks for model evaluation. All experiments were conducted on the NVIDIA RTX 8000 GPUs.

% \begin{table}[hbtp]
%     \centering
%     \resizebox{\linewidth}{!}{%
%     \begin{tabular}{c|c|c|c|c|c|c}
%     \toprule
%     \bfseries Dataset & \bfseries VLM & \multicolumn{2}{c|}{\bfseries Unique Features} & \multicolumn{2}{c|}{\bfseries Avg. Features per Class} & \bfseries Overlap Features \\
%     % \cline{3-6}
%     & & Train & Test & Train & Test & \\ 
%     \midrule
%     miniImageNet & vit-gpt2 & 1908 & 1111 & 191.28 & 190.40 & 931 \\
%     miniImageNet & blip & 4097 & 2032 & 236.16 & 254.40 & 1288 \\
%     tieredImageNet & vit-gpt2 & 3832 & 2532 & 249.94 & 230.94 & 1994 \\
%     tieredImageNet & blip & 12188 & 6710 & 321.47 & 310.40 & 3725 \\
%     CUB-200-2011 & vit-gpt2 & 211 & 159 & 25.38 & 25.78 & 136 \\
%     CUB-200-2011 & blip & 377 & 247 & 31.53 & 29.74 & 203 \\
%     \bottomrule
%     \end{tabular}%
%     }
%     \caption{Number of detected features in train and test datasets with different feature extractors.}\label{tab:feat-statistics}
% \end{table}

\begin{figure}[t]
    \centering
    \includegraphics[width=\linewidth]{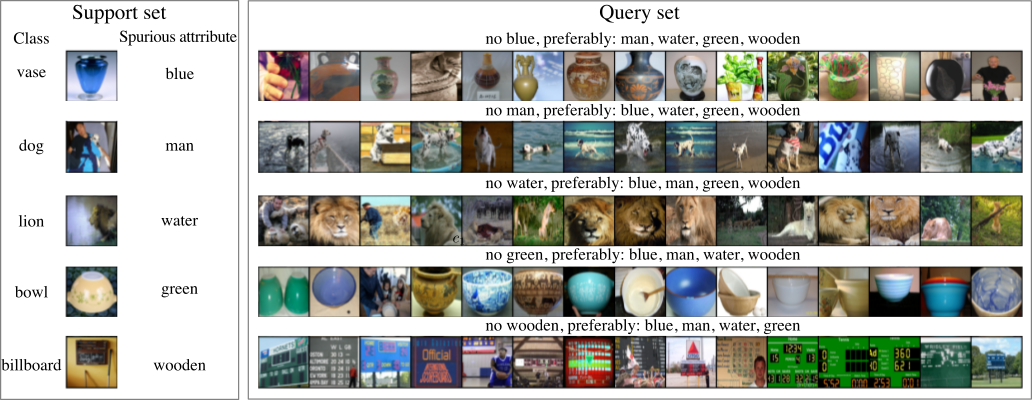}
    \caption{A 5-way 1-shot task constructed by our inter-class attribute-based sample selection using samples from the miniImageNet dataset. Note that due to the limited capacity of a VLM, the attributes may not well align with human understandings.}
    \label{fig:task-example}
\end{figure}

\subsection{Visualization of a Constructed Task}
We show a 5-way 1-shot task constructed by FewSTAB in Fig. \ref{fig:task-example}. Each class in the support set correlates with a unique spurious attribute. The query samples of a class do not have the spurious attribute correlated with the class and some of them have spurious attributes associated with other classes in the support set. For example, the query samples of the class \texttt{lion} do not have the spurious attribute \texttt{water},  and some of them have spurious attributes from other classes in the support set, such as \texttt{man} and \texttt{green}. 
%Similarly, the query samples of class \texttt{dog} do not have the spurious attribute \texttt{man}, and some of them have the spurious attribute \texttt{water} from the class \texttt{lion} in the support set. 
FewSTAB introduces biased attributes in the task so that query samples can be easily misclassified as other classes by a few-shot classifier that relies on the spurious correlations in the support set.

\renewcommand{\minval}{5}
\renewcommand{\maxval}{80}
\renewcommand{\opacity}{25}
\begin{table}[t]
\centering
\caption{Comparison between wAcc-R and wAcc-A with 95\% confidence interval on the miniImageNet, tieredImageNet, and CUB datasets. Numbers in the Shot column indicate that the models are both trained (if applicable) and tested on 5-way 1- or 5-shot tasks.  Darker colors indicate higher values.}
\resizebox{0.8\linewidth}{!}{%
\begin{tabular}{c|l|c@{\hskip 0.2cm}c@{\hskip 0.2cm}c@{\hskip 0.2cm}c@{\hskip 0.2cm}c@{\hskip 0.2cm}c}
\hline
\multirow{2}{*}{Shot} & \multicolumn{1}{c|}{\multirow{2}{*}{Method}}  & \multicolumn{2}{c}{miniImageNet} & \multicolumn{2}{c}{tieredImageNet} & \multicolumn{2}{c}{CUB-200} \\ \cline{3-8} 
&  & wAcc-A ($\uparrow$) & wAcc-R ($\uparrow$) & wAcc-A ($\uparrow$) & wAcc-R ($\uparrow$) & wAcc-A ($\uparrow$) & wAcc-R ($\uparrow$) \\ \hline
\multirow{10}{*}{1} & ANIL & \gradient{10.38}$\pm$0.30 & \gradient{14.36}$\pm$0.33 & \gradient{11.21}$\pm$0.30 & \gradient{15.63}$\pm$0.36 & \gradient{13.78}$\pm$0.40 & \gradient{16.94}$\pm$0.43 \\
& LEO & \gradient{14.26}$\pm$0.46 & \gradient{21.35}$\pm$0.54 & \gradient{16.00}$\pm$0.55 & \gradient{29.63}$\pm$0.71 & \gradient{28.29}$\pm$0.80 & \gradient{40.22}$\pm$0.87 \\
& BOIL & \gradient{12.48}$\pm$0.23 & \gradient{14.93}$\pm$0.24 & \gradient{12.27}$\pm$0.21 & \gradient{14.13}$\pm$0.23 & \gradient{19.15}$\pm$0.29 & \gradient{22.50}$\pm$0.29 \\
& ProtoNet & \gradient{14.03}$\pm$0.49 & \gradient{21.96}$\pm$0.58 & \gradient{14.50}$\pm$0.50 & \gradient{27.13}$\pm$0.69 & \gradient{34.62}$\pm$0.85 & \gradient{46.61}$\pm$0.89 \\
& DN4 & \gradient{12.37}$\pm$0.46 & \gradient{19.28}$\pm$0.56 & \gradient{11.99}$\pm$0.47 & \gradient{23.62}$\pm$0.65 & \gradient{35.22}$\pm$0.86 & \gradient{47.26}$\pm$0.88 \\
& R2D2 & \gradient{18.05}$\pm$0.53 & \gradient{26.50}$\pm$0.60 & \gradient{16.41}$\pm$0.54 & \gradient{30.41}$\pm$0.71 & \gradient{36.70}$\pm$0.90 & \gradient{48.82}$\pm$0.88 \\
& CAN & \gradient{17.37}$\pm$0.53 & \gradient{25.96}$\pm$0.60 & \gradient{18.84}$\pm$0.60 & \gradient{36.29}$\pm$0.78 & \gradient{22.74}$\pm$0.72 & \gradient{31.95}$\pm$0.78 \\
& RENet & \gradient{19.10}$\pm$0.57 & \gradient{28.85}$\pm$0.65 & \gradient{18.83}$\pm$0.61 & \gradient{35.70}$\pm$0.78 & \gradient{32.43}$\pm$0.81 & \gradient{43.98}$\pm$0.81 \\
& Baseline++ & \gradient{15.30}$\pm$0.48 & \gradient{23.18}$\pm$0.56 & \gradient{17.51}$\pm$0.54 & \gradient{31.62}$\pm$0.71 & \gradient{9.17}$\pm$0.47 & \gradient{14.59}$\pm$0.58 \\
& RFS & \gradient{18.00}$\pm$0.53 & \gradient{27.12}$\pm$0.61 & \gradient{18.35}$\pm$0.60 & \gradient{35.24}$\pm$0.77 & \gradient{32.45}$\pm$0.80 & \gradient{44.49}$\pm$0.82 \\ \hline
\multirow{10}{*}{5} & ANIL& \gradient{14.83}$\pm$0.40 & \gradient{25.37}$\pm$0.52 & \gradient{13.72}$\pm$0.39 & \gradient{30.60}$\pm$0.51 & \gradient{31.63}$\pm$0.55 & \gradient{45.47}$\pm$0.53 \\
& LEO& \gradient{26.31}$\pm$0.59 & \gradient{41.33}$\pm$0.59 & \gradient{29.49}$\pm$0.72 & \gradient{57.22}$\pm$0.72 & \gradient{46.62}$\pm$0.82 & \gradient{59.76}$\pm$0.77 \\
& BOIL& \gradient{13.09}$\pm$0.22 & \gradient{15.21}$\pm$0.23 & \gradient{14.90}$\pm$0.22 & \gradient{18.55}$\pm$0.24 & \gradient{19.17}$\pm$0.28 & \gradient{21.33}$\pm$0.27 \\
& ProtoNet& \gradient{32.07}$\pm$0.58 & \gradient{51.95}$\pm$0.52 & \gradient{30.95}$\pm$0.70 & \gradient{62.53}$\pm$0.62 & \gradient{60.06}$\pm$0.74 & \gradient{75.68}$\pm$0.50 \\
& DN4& \gradient{27.60}$\pm$0.58 & \gradient{42.68}$\pm$0.62 & \gradient{16.07}$\pm$0.62 & \gradient{40.63}$\pm$0.81 & \gradient{59.25}$\pm$0.77 & \gradient{73.58}$\pm$0.56 \\
& R2D2& \gradient{35.37}$\pm$0.59 & \gradient{50.84}$\pm$0.55 & \gradient{31.12}$\pm$0.72 & \gradient{61.08}$\pm$0.65 & \gradient{58.66}$\pm$0.82 & \gradient{75.20}$\pm$0.54 \\
& CAN& \gradient{36.44}$\pm$0.65 & \gradient{54.23}$\pm$0.55 & \gradient{31.17}$\pm$0.76 & \gradient{64.19}$\pm$0.62 & \gradient{41.31}$\pm$0.74 & \gradient{61.61}$\pm$0.58 \\
& RENet& \gradient{36.19}$\pm$0.63 & \gradient{56.52}$\pm$0.58 & \gradient{30.27}$\pm$0.76 & \gradient{63.49}$\pm$0.64 & \gradient{52.93}$\pm$0.82 & \gradient{71.82}$\pm$0.56 \\
& Baseline++& \gradient{29.52}$\pm$0.57 & \gradient{44.94}$\pm$0.57 & \gradient{30.01}$\pm$0.72 & \gradient{59.06}$\pm$0.67 & \gradient{16.86}$\pm$0.52 & \gradient{29.84}$\pm$0.69 \\
& RFS& \gradient{36.85}$\pm$0.64 & \gradient{55.66}$\pm$0.55 & \gradient{31.15}$\pm$0.76 & \gradient{62.71}$\pm$0.67 & \gradient{54.98}$\pm$0.81 & \gradient{74.33}$\pm$0.53 \\ \bottomrule
\end{tabular}%
}
\label{tab:worst-classification}
%\vspace{-4mm}
\end{table}

\begin{figure}[t]
    \centering
    \includegraphics[width=\linewidth]{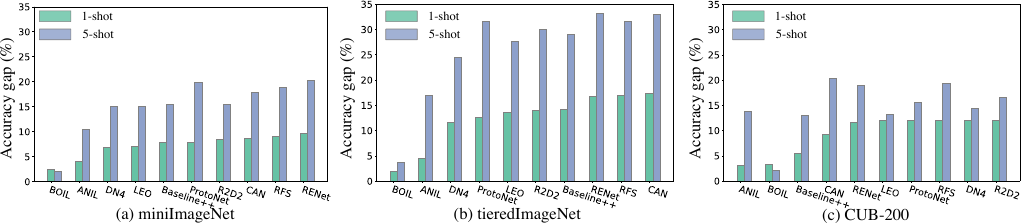}
    \caption{Accuracy gaps (wAcc-R minus wAcc-A) on the 5-way 1-shot and 5-way 5-shot tasks from the (a) miniImageNet, (b) tieredImageNet, and (c) CUB-200 datasets. }
    \label{fig:sa_r_gap_cmp}
\end{figure}

\subsection{Effectiveness of FewSTAB}\label{sec:effectiveness-fewstab}

\noindent \textbf{FewSTAB can effectively reveal spurious bias in few-shot classifiers.}  
We show in Table \ref{tab:worst-classification} the wAcc (Eq. \eqref{eq:worst-accuracy}) on 5-way 1/5-shot test tasks  that 
are randomly
sampled (wAcc-R) and are constructed with FewSTAB (wAcc-A), respectively. FewSTAB generates FSC test tasks only based on the class-attribute correlations in data. In each test setting, the FSC methods in Table \ref{tab:worst-classification} are evaluated with the same FSC tasks. 
  We observe that wAcc-A is consistently lower than wAcc-R on the three datasets and on two test-shot numbers, showing that FewSTAB 
is more effective than the standard evaluation procedure (random task construction) in exhibiting the spurious bias in various few-shot classifiers. We additionally show that FewSTAB also works on the most recent FSC methods and can reflect the improvement made to mitigate spurious bias (see Appendix).

\begin{wraptable}{R}{0.4\linewidth}
\centering
\caption{Spearman's rank correlations between wAcc-A and wAcc-R in Table \ref{tab:worst-classification}.}
\resizebox{0.8\linewidth}{!}{%
\begin{tabular}{lll}
\toprule
Dataset        & 1-shot & 5-shot \\ \hline
miniImageNet   & 0.96   & 0.95   \\
tieredImageNet & 0.96   & 0.90   \\
CUB-200        & 1.00   & 0.94   \\ \bottomrule
\end{tabular}%
}
\label{tab:spearman-corr-accw}
\end{wraptable}

% \begin{table}[t]
% \centering
% \caption{Spearman's rank correlations between wAcc-A and wAcc-R in \cref{tab:worst-classification}.}
% \resizebox{0.4\linewidth}{!}{%
% \begin{tabular}{cccc}
% \hline
% Shot & miniImageNet & tieredImageNet & CUB-200 \\ \hline
% 1    & 0.96         & 0.96           & 1.00    \\
% 5    & 0.95         & 0.90           & 0.94    \\ \hline
% \end{tabular}%
% }
% \label{tab:spearman-corr-accw}
% \end{table}

\noindent \textbf{FewSTAB reveals new robustness patterns among FSC methods.} In Table \ref{tab:spearman-corr-accw}, we calculate the 
Spearman's rank correlation coefficients \cite{myers2004spearman} between
the values of wAcc-R and wAcc-A from Table \ref{tab:worst-classification}. The coefficients are bounded  from 0 to 1, with larger values indicating that the ranks of FSC methods based on wAcc-R are more similar to those based on wAcc-A. In the 1-shot setting, 
it is not effective to control the spurious correlations since we only have one sample per class in the support set. Hence, the coefficients are large, and the ranks based on wAcc-A are similar to those based on wAcc-R. In the 5-shot cases, we have more samples to  demonstrate the spurious correlations. The coefficients become smaller, \ie, the ranks based on wAcc-A show different trends from those based on wAcc-R. In this case, FewSTAB reveals new information on FSC methods' varied degrees of robustness to spurious bias.

\noindent \textbf{FewSTAB can benchmark spurious bias in varied degrees.} As shown in Fig. \ref{fig:sa_r_gap_cmp}, the accuracy gap, defined as wAcc-R minus wAcc-A, in general, becomes larger when we switch from 5-way 1-shot to 5-way 5-shot tasks. Compared with the random task construction, FewSTAB creates more challenging tasks in the 5-shot case for demonstrating spurious bias in few-shot classifiers. In other words, with a higher shot value in the constructed test tasks, FewSTAB aims to benchmark spurious bias in a higher degree.

% Moreover, compared with the other two datasets,  accuracy gaps are generally larger on the tieredImageNet dataset which has richer classes and more distinct features (Table \ref{tab:feat-statistics}) that can be exploited by FewSTAB.  

% \begin{table}[h]
% \centering
% %\resizebox{0.8\linewidth}{!}{%
% \begin{tabular}{cccc}
% \hline
% Shot & miniImageNet & tieredImageNet & CUB-200 \\ \hline
% 1 & 0.96 & 0.96 & 1.00 \\
% 5 & 0.95 & 0.90 & 0.94 \\ \hline
% \end{tabular}%
% %}
% \caption{Spearman's rank correlation coefficients between wAcc-A and wAcc-R for the models tested in Table \ref{tab:worst-classification}.}
% \label{tab:spearman-corr-accw}
% \end{table}

% \begin{table}[h]
% \centering
% \resizebox{0.6\linewidth}{!}{%
% \begin{tabular}{lcc}
% \hline
% Dataset & 1-shot & 5-shot \\ \hline
% miniImageNet & 0.96 & 0.95 \\
% tieredImageNet & 0.96 & 0.90 \\
% CUB-200 & 1.00 & 0.94 \\ \hline
% \end{tabular}%
% }
% \caption{Spearman's rank correlation coefficients between wAcc-A and wAcc-R for the models tested in Table \ref{tab:worst-classification}.}
% \label{tab:spearman-corr-accw}
% \end{table}

\subsection{A New Dimension of Evaluation and a New Design Guideline}
\begin{wrapfigure}{r}{0.4\linewidth}
    \centering
    \includegraphics[width=0.9\linewidth]{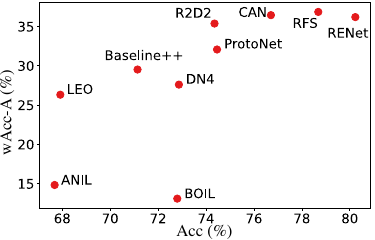}
    \caption{Acc versus wAcc-A of the ten FSC methods tested on 5-way 5-shot tasks from miniImageNet. }
    \label{fig:avg-accw-cmp}
\end{wrapfigure}
FewSTAB creates a new dimension of evaluation on the robustness to spurious bias. 
We demonstrate this
with a scatter plot (Fig. \ref{fig:avg-accw-cmp}) of Acc  (Eq. \eqref{eq:standard-accuracy})
   and wAcc-A of the ten few-shot classifiers. FewSTAB offers new information regarding different few-shot classifiers' robustness to spurious bias as we observe that Acc does not well correlate with wAcc-A.
A high wAcc-A indicates that the classifier is robust to spurious bias, while a high Acc indicates that the classifier can correctly predict most of the samples. With the scatter plot, we can view tradeoffs between the two metrics on existing few-shot classifiers. A desirable few-shot classifier should appear in the top-right corner of the plot.

\begin{wrapfigure}{R}{0.5\linewidth}
    \centering
    %\vspace{-8mm}
    \includegraphics[width=\linewidth]{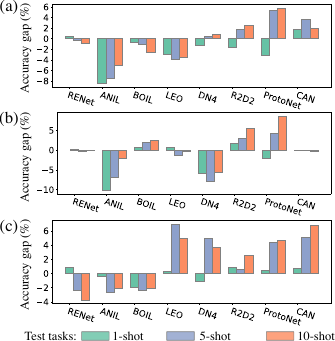}
    \caption{Accuracy gaps of few-shot classifiers tested on 1-shot, 5-shot, and 10-shot tasks constructed from (a) miniImageNet, (b) tieredImageNet, and (c) CUB-200 datasets.}
    \label{fig:effectiveness-train-shot}
\end{wrapfigure}

\subsection{FewSTAB Enables Designs for Varied Degrees of Robustness}
As demonstrated in Section \ref{sec:effectiveness-fewstab}, FewSTAB can benchmark spurious bias in varied degrees, which in turn enables practitioners to design robust few-shot classifiers targeted for different degrees of robustness to spurious bias. The reason for differentiating designs for varied degrees of robustness is that the same design choice may not work under different robustness requirements. For example, increasing shot number in training tasks is a common strategy for improving the few-shot generalization of meta-learning based methods. We trained few-shot classifiers with 5-way 5-shot and 5-way 1-shot training tasks randomly sampled from $\mathcal{D}_{train}$, respectively. We then calculated the \textit{accuracy gap defined as the wAcc-A of a model trained on 5-shot tasks minus the wAcc-A of the same model trained on 1-shot tasks}. A positive and large accuracy gap indicates that this strategy is effective in improving the model's robustness to spurious bias. 
In Fig. \ref{fig:effectiveness-train-shot}, on each of the three datasets, we give results of the eight meta-learning based FSC methods on the 5-way 1-, 5-, and 10-shot FewSTAB tasks which are used to demonstrate the strategy's robustness to increased degrees of spurious bias. This strategy does not work consistently under different test shots. For example, in Fig. \ref{fig:effectiveness-train-shot}(a) this strategy with CAN only works the best on the 5-way 5-shot FewSTAB tasks.

\subsection{Ablation Studies}\label{sec:ablation}
\noindent\textbf{Techniques used in FewSTAB.} We analyze how different sample selection methods affect the effectiveness of FewSTAB in Table \ref{tab:fsc-task-construction-methods}. With only intra-class attribute-based sample selection, we randomly select query samples from Eq. \eqref{eq:intra-class-attack}. For inter-class attribute-based sample selection and intra-class attribute-based sample selection (automatically included by Eq. \eqref{eq:cls-attack-set}), we randomly select query samples from Eq. \eqref{eq:multi-cls-attack-set}. FewSTAB uses all the techniques in Table \ref{tab:fsc-task-construction-methods}.  We define accuracy drop as wAcc-R minus wAcc-A, and we use the drop averaged over the ten FSC methods tested on 5-way 5-shot tasks from the miniImageNet dataset as our metric.
A larger average drop indicates that the corresponding sample selection method is more effective in reflecting the spurious bias in few-shot classifiers. We observe that all proposed techniques are effective and the inter-class attribute-based sample selection is the most effective method. 

\begin{table}[t]
\centering
   \begin{minipage}[t]{.5\linewidth}
   \centering
   \caption{Comparison between different techniques used by FewSTAB for constructing FSC tasks.}
   \resizebox{0.8\linewidth}{!}{%
       \begin{tabular}{cccc}
\toprule
\multicolumn{2}{c}{\begin{tabular}[c]{@{}c@{}}Attribute-based \\sample selection\end{tabular}} & \multirow{2}{*}{\begin{tabular}[c]{@{}c@{}}Query sample \\ selection\end{tabular}} & \multirow{2}{*}{Avg. drop (\%)} \\ \cmidrule{1-2}
Intra-class & Inter-class &  &  \\ \midrule
$\checkmark$ &  &  & 5.13 \\
$\checkmark$ & $\checkmark$ &  & 13.30 \\
$\checkmark$ & $\checkmark$ & $\checkmark$ & 15.05 \\ \bottomrule
\end{tabular}%
}
\label{tab:fsc-task-construction-methods}
   \end{minipage}\hfill
   \begin{minipage}[t]{.43\linewidth}
   \centering
   \caption{Spearman's rank correlation coefficients between wAcc-A obtained using ViT-GPT2 and BLIP.}
\resizebox{0.68\linewidth}{!}{%
       \begin{tabular}{lcc}
\toprule
Dataset & 1-shot & 5-shot \\ \midrule
miniImageNet & 0.98 & 1.00 \\
tieredImageNet & 1.00 & 0.99 \\
CUB-200 & 1.00 & 0.98 \\ \bottomrule
\end{tabular}%
}
\label{tab:spearman-corr-vlm}
   \end{minipage}
\end{table}

% \begin{table}[htbp]
% \centering
% %\resizebox{0.6\linewidth}{!}{%
% \begin{tabular}{cccc}
% \toprule
% \multicolumn{2}{c}{attribute bias} & \multirow{2}{*}{\begin{tabular}[c]{@{}c@{}}Query sample \\ selection\end{tabular}} & \multirow{2}{*}{Avg. drop (\%)} \\ \cmidrule{1-2}
% Intra-class & Inter-class &  &  \\ \midrule
% $\checkmark$ &  &  & 5.13 \\
% $\checkmark$ & $\checkmark$ &  & 13.30 \\
% $\checkmark$ & $\checkmark$ & $\checkmark$ & 15.05 \\ \bottomrule
% \end{tabular}%
% %}
% \caption{Comparison between different techniques used by FewSTAB for constructing FSC tasks.}
% \label{tab:fsc-task-construction-methods}
% \end{table}
%FewSTAB pushes the evaluation the robustness of few-shot classifiers closer to their performance limits.

\noindent\textbf{Choice of VLMs.}
Although our main results are based on the pre-trained ViT-GPT2 model~\cite{nlp_connect_2022}, we show in Table \ref{tab:spearman-corr-vlm} that when switching to a different VLM, \ie, BLIP~\cite{li2022blip}, the relative ranks of different few-shot classifiers based on wAcc-A still hold with high correlations. In other words, FewSTAB is robust to different choices of VLMs.

\noindent\textbf{Detection accuracy of VLMs.} A VLM may miss some attributes due to its limited capacity, resulting in a small detection accuracy. However, the detection accuracy of a VLM has little impact on our framework.
To demonstrate this, we adopt a cross-validation strategy, \ie, we use the outputs from one VLM as the ground truth to evaluate those from another VLM, since assessing the detection accuracy of a VLM typically requires labor-intensive human labeling. On the CUB-200 dataset, we observe that the detection accuracy of ViT-GPT2 based on the BLIP's outputs is 70.12\%, while the detection accuracy of BLIP based on the ViT-GPT2's outputs is 59.28\%. Although the two VLMs differ significantly in the detected attributes, our framework shows almost consistent rankings of the evaluated FSC methods (Table \ref{tab:spearman-corr-vlm}).
% 
% We do not include RFS and Baseline++ since they are pretraining methods, and their accuracy gaps are zero. We observe that this simple strategy only works for some methods on certain datasets. For example, ANIL has negative gaps in all cases. For metric-based methods that rely on meta-learned embeddings, such as ProtoNet and R2D2, increasing training shots helps learn robust representations and improves robustness against spurious correlations. For CAN, which adopts attention mechanisms for classification, increasing training shots is not effective on the tieredImageNet dataset, which has the most classes. However, it is beneficial for CAN on the fine-grained CUB-200 dataset. In the future, it is worth exploring new training methods that work for most of the methods.

\noindent Additional results are presented in Appendix.

\section{Conclusion}
In this paper, we proposed a systematic and rigorous benchmark framework called FewSTAB for evaluating the robustness of few-shot classifiers to spurious bias. FewSTAB adopts attribute-based sample selection strategies to construct FSC test tasks with biased attributes so that the reliance on spurious correlations can be effectively revealed.  FewSTAB can automatically benchmark spurious bias in few-shot classifiers on any existing test data thanks to its use of a pre-trained VLM for automated attribute detection. With FewSTAB, we provided a new dimension of evaluation on the robustness of few-shot image classifiers to spurious bias and a new design guideline for building robust few-shot classifiers. FewSTAB can reveal and enable designs for varied degrees of robustness to spurious bias. We hope FewSTAB will inspire new developments on designing robust few-shot classifiers.

\section*{Acknowledgments}
This work is supported in part by the US National Science Foundation under grants 2313865, 2217071, 2213700, 2106913, 2008208, 1955151.
{
    \bibliographystyle{splncs04}
    \bibliography{main}
}

\clearpage
\appendix
\setcounter{page}{1}
\setcounter{section}{0}
\setcounter{table}{0}
\setcounter{figure}{0}
\renewcommand{\thesection}{\Alph{section}}
\renewcommand{\thetable}{S\arabic{table}}
\renewcommand{\thefigure}{S\arabic{figure}}

\section*{Appendix}
The appendix is organized as follows:
we introduce the ten FSC algorithms adopted in the paper in \cref{sec:fsc-algorithms}. Then, we give the details of the evaluation metrics used in the main paper in \cref{sec:eval-metrics}. In \cref{sec:exp-setting-details}, we show statistics of the datasets used in this paper along with detailed training settings. In \cref{sec:additional-ablation-studies}, we analyze different methods for constructing the support and query sets in a FewSTAB task (\cref{sec:appendix-ablation-studies}), show the scatter plots of wAcc-A versus Acc from all the training settings (\cref{sec:appendix-scatter-plots}), present more results on the effectiveness of FewSTAB (\cref{sec:effectiveness-more-results}), and demonstrate the robustness of FewSTAB with different VLMs (\cref{sec:appendix-different-vlms}). Finally, we give more examples of the tasks constructed by FewSTAB in \cref{sec:more-visualizations}.

\section{Few-Shot Classification Algorithms}\label{sec:fsc-algorithms}

    \noindent \textbf{ANIL (Almost No Inner Loop)} \cite{Raghu2020Rapid}: ANIL is an optimization-based meta-learning method and follows a similar optimization procedure to MAML \cite{finn2017model} whose few-shot adaptation algorithm $\mathcal{O}$ is to update the whole model using gradient descent with a few learning samples.  ANIL does not update the whole model and instead only updates the classifier in the last layer.
    
    \noindent \textbf{BOIL (Body Only update in Inner Loop)} \cite{oh2020boil}: BOIL is another optimization-based meta-learning method. Its adaptation algorithm $\mathcal{O}$ freezes the update of the classifier and only updates the embedding backbone.
    
    \noindent \textbf{LEO (Latent Embedding Optimization)} \cite{rusu2018metalearning}:  LEO is similar to MAML. But instead of directly optimizing high-dimensional model parameters, its adaptation algorithm $\mathcal{O}$  learns a generative distribution of model parameters and optimizes the model parameters in a low-dimensional latent space.
    
    \noindent \textbf{ProtoNet (Prototypical Networks)} \cite{Snell2017Prototypical}:   ProtoNet is a metric-based meta-learning method. Its adaptation algorithm $\mathcal{O}$ first  calculates a prototype representation for each class as the mean vector of each support class, and then uses a nearest-neighbor classifier created with the class prototypes and the Euclidean distance function to predict a query image. 

    \noindent \textbf{DN4 (Deep Nearest Neighbor Neural Network)} \cite{li2019revisiting}: DN4 is a metric-based meta-learning method, which does not use attributes after pooling for classification. Instead, DN4 uses the local attributes before pooling and employs a local descriptor based image-to-class measure for classification.

    \noindent  \textbf{R2D2 (Ridge Regression Differentiable Discriminator)} \cite{bertinetto2018metalearning}: R2D2 is a metric-based meta-learning method and adopts ridge regression as the few-shot adaptation algorithm $\mathcal{O}$. The advantage of R2D2 is that ridge regression
    enjoys a closed-form solution and can learn efficiently with a few training samples. 

    \noindent \textbf{CAN (Cross Attention Network) }\cite{hou2019cross}: CAN is a metric-based meta-learning method and calculates the cross attention between each pair of class and query features so as to exploit and learn discriminative features for predictions.

    \noindent \textbf{RENet (Relational Embedding Network)} \cite{kang2021relational}: RENet is a  metric-based meta-learning method. It uses a self-correlational representation module and a  cross-correlational attention module to learn relational patterns within and between images, respectively.

    \noindent \textbf{RFS (Rethinking Few-Shot)} \cite{tian2020rethinking}: RFS is a transfer learning method. It first trains an embedding network using base classes. Then, instead of fine-tuning the last fully-connected classification layer, it learns a new logistic regression classifier with L2-normalized feature vectors from a few samples of novel classes.

    \noindent \textbf{Baseline++} \cite{chen2018a}: Baseline++ is a transfer learning method. It first pretrains an embedding network using samples from base classes. Then, it fine-tunes the last fully-connected layer with a few samples of novel classes but replaces the standard inner product  with a cosine distance between input features and the weight vectors of the layer.

\section{Evaluation Metrics}\label{sec:eval-metrics}
\noindent\textbf{Standard accuracy (Acc):} Acc measures on average how a few-shot classifier generalizes to different tasks with novel classes not seen before. We define Acc as follows,
\begin{align}\label{eq:standard-accuracy-appendix}
\text{Acc}=\frac{1}{N_T}\sum_{t=1}^{N_T}\sum_{c=1}^CM_c(\mathcal{T}_t;f_\theta, \mathcal{O}),
\end{align}
where $N_T$ is the number of test tasks, $C$ is the number of classes per task, $\mathcal{T}_t$ is the $t$-th $C$-way $N_S$-shot task with $N_Q$ query samples per class, $f_\theta$ is a few-shot classifier, $\mathcal{O}$ is the few-shot adaptation algorithm associated with $f_\theta$, $M_c$ denotes the classification accuracy on the \textit{query samples} of the class $c$. This metric is used in \cref{fig:avg-accw-cmp}.

\noindent\textbf{Class-wise worst classification accuracy (wAcc):} wAcc characterizes the performance limit of $f_\theta$  in learning novel classes, and we calculate  wAcc as the average of the smallest per-class classification accuracy on query samples over $N_T$ tasks, i.e.,
\begin{align}\label{eq:worst-accuracy-appendix}
\text{wAcc}=\frac{1}{N_T}\sum_{t=1}^{N_T}\min_{c=1,\ldots,C}M_c(\mathcal{T}_t;f_\theta, \mathcal{O}).
\end{align}
Depending on what kinds of tasks are used for evaluation, we have the following two types of wAcc:
\begin{itemize}
    \item \textbf{wAcc-R:} If the test tasks are randomly sampled in Eq. \eqref{eq:worst-accuracy-appendix}, then we get wAcc-R on $N_T$ randomly sampled tasks. This metric is used in \cref{tab:worst-classification} as a baseline for highlighting the effectiveness of our FewSTAB in revealing the spurious bias in few-shot classifiers.

    \item\textbf{wAcc-A:} If the $N_T$ test tasks in Eq. \eqref{eq:worst-accuracy-appendix} are constructed by our FewSTAB, then we get wAcc-A, which characterizes the robustness of a few-shot classifier to spurious bias. This metric is the main metric used in the experiments.
\end{itemize}

\noindent\textbf{Accuracy gap between wAcc-R and wAcc-A:} We obtain the wAcc-R and wAcc-A of a model by testing it with tasks randomly sampled and with tasks constructed by FewSTAB, respectively. The accuracy gap is calculated as the wAcc-R minus the wAcc-A. A large gap indicates the effectiveness of FewSTAB in revealing the robustness of a few-shot classifier to spurious bias. This metric is used in \cref{fig:sa_r_gap_cmp} and \cref{tab:fsc-task-construction-methods}.

 \noindent\textbf{Accuracy gap between wAcc-A of models trained with different shots:}
 We train a few-shot classifier with $C$-way (e.g. 5-way) 5-shot and 1-shot training tasks from $\mathcal{D}_{train}$, respectively. Then, we test the obtained two classifiers with the \textit{same} tasks created by FewSTAB and calculate the accuracy gap as the wAcc-A of the model trained with 5-shot tasks minus the wAcc-A of the model trained with 1-shot tasks. A large accuracy gap indicates that increasing training shots can improve a few-shot classifier's robustness to spurious bias. This metric is used in \cref{fig:effectiveness-train-shot}.

\begin{table}[thbp]
\centering
\caption{Numbers of classes along with numbers of samples (in parentheses) in each split of the three datasets.}
%\resizebox{0.5\linewidth}{!}{%
\begin{tabular}{cccc}
\hline
Split & miniImageNet & tieredImageNet & CUB-200 \\ \hline
$\mathcal{D}_{train}$ & 64 (38.4k) & 351 (448.7k) & 130 (7.6k) \\
$\mathcal{D}_{val}$ & 16 (9.6k) & 97 (124.3k) & 20 (1.2k) \\
$\mathcal{D}_{test}$ & 20 (12k) & 160 (206.2k) & 50 (3.0k) \\ \hline
\end{tabular}%
%}
\label{tab:dataset}
\end{table}

\begin{table*}[t]
\centering
\caption{Training configurations and hyperparameters for training on the miniImageNet dataset. ``-'' denotes not applicable.}
\resizebox{\linewidth}{!}{%
\begin{tabular}{ccclcccc}
\hline
Method & Mode & Learning rate & LR scheduler & Optimizer & Epochs & Training episodes & Episode size \\ \hline
\multirow{2}{*}{ANIL} & T (5w1s) & 0.001 & - & Adam & 100 & 2000 & 4 \\
 & T (5w5s) & 0.001 & - & Adam & 100 & 2000 & 4 \\ \hline
\multirow{2}{*}{LEO} & T (5w1s) & 0.0005 & CosineAnnealingLR & Adam & 100 & 2000 & 1 \\
 & T (5w5s) & 0.001 & CosineAnnealingLR & Adam & 100 & 2000 & 1 \\ \hline
\multirow{2}{*}{BOIL} & T (5w1s) & 0.0006 & - & Adam & 100 & 2000 & 4 \\
 & T (5w5s) & 0.0006 & - & Adam & 100 & 2000 & 4 \\ \hline
\multirow{2}{*}{ProtoNet} & T (5w1s) & 0.001 & StepLR(20, 0.5) & Adam & 100 & 200 & 1 \\
 & T (5w5s) & 0.001 & StepLR(20, 0.5) & Adam & 100 & 2000 & 1 \\ \hline
\multirow{2}{*}{DN4} & T (5w1s) & 0.001 & StepLR(50, 0.5) & Adam & 100 & 2000 & 1 \\
 & T (5w5s) & 0.001 & StepLR(50, 0.5) & Adam & 100 & 2000 & 1 \\ \hline
\multirow{2}{*}{R2D2} & T (5w1s) & 0.1 & CosineAnnealingLR & SGD & 100 & 2000 & 4 \\
 & T (5w5s) & 0.1 & CosineAnnealingLR & SGD & 100 & 2000 & 4 \\ \hline
\multirow{2}{*}{CAN} & T (5w1s) & 0.1 & CosineAnnealingLR & SGD & 100 & 2000 & 8 \\
 & T (5w5s) & 0.1 & CosineAnnealingLR & SGD & 100 & 2000 & 4 \\ \hline
\multirow{2}{*}{RENet} & T (5w1s) & 0.1 & CosineAnnealingLR & SGD & 100 & 300 & 1 \\
 & T (5w5s) & 0.1 & CosineAnnealingLR & SGD & 100 & 300 & 1 \\ \hline
{Baseline++} & B (128) & 0.01 & CosineAnnealingLR & SGD & 100 & - & - \\ \hline
{RFS} & B (64) & 0.05 & MultiStepLR([60, 80], 0.1) & SGD & 100 & - & - \\ \hline
\end{tabular}%
}
\label{tab:train-mini}
\end{table*}

\begin{table*}[htbp]
\centering
\caption{Training configurations and hyperparameters for training on the tieredImageNet dataset. ``-'' denotes not applicable.}
\resizebox{\linewidth}{!}{%
\begin{tabular}{ccclcccc}
\hline
Method & Mode & Learning rate & LR scheduler & Optimizer & Epochs & Training episodes & Episode size \\ \hline
\multirow{2}{*}{ANIL} & T (5w1s) & 0.001 & - & Adam & 100 & 5000 & 4 \\
 & T (5w5s) & 0.001 & - & Adam & 100 & 5000 & 4 \\ \hline
\multirow{2}{*}{LEO} & T (5w1s) & 0.0005 & CosineAnnealingLR & Adam & 100 & 5000 & 1 \\
 & T (5w5s) & 0.001 & CosineAnnealingLR & Adam & 100 & 5000 & 1 \\ \hline
\multirow{2}{*}{BOIL} & T (5w1s) & 0.0006 & - & Adam & 100 & 5000 & 4 \\
 & T (5w5s) & 0.0006 & - & Adam & 100 & 5000 & 4 \\ \hline
\multirow{2}{*}{ProtoNet} & T (5w1s) & 0.001 & StepLR(20, 0.5) & Adam & 100 & 5000 & 1 \\
 & T (5w5s) & 0.001 & StepLR(20, 0.5) & Adam & 100 & 5000 & 1 \\ \hline
\multirow{2}{*}{DN4} & T (5w1s) & 0.001 & StepLR(50, 0.5) & Adam & 200 & 5000 & 1 \\
 & T (5w5s) & 0.001 & StepLR(50, 0.5) & Adam & 200 & 5000 & 1 \\ \hline
\multirow{2}{*}{R2D2} & T (5w1s) & 0.1 & CosineAnnealingLR & SGD & 100 & 5000 & 4 \\
 & T (5w5s) & 0.1 & CosineAnnealingLR & SGD & 100 & 5000 & 4 \\ \hline
\multirow{2}{*}{CAN} & T (5w1s) & 0.1 & CosineAnnealingLR & SGD & 100 & 2000 & 4 \\
 & T (5w5s) & 0.1 & CosineAnnealingLR & SGD & 100 & 2000 & 4 \\ \hline
\multirow{2}{*}{RENet} & T (5w1s) & 0.1 & MultiStepLR([60, 80], 0.05) & SGD & 100 & 1752 & 1 \\
 & T (5w5s) & 0.1 & MultiStepLR([40, 50], 0.05) & SGD & 60 & 1752 & 1 \\ \hline
{Baseline++} & B (128) & 0.01 & CosineAnnealingLR & SGD & 100 & - & - \\ \hline
{RFS} & B (128) & 0.1 & CosineAnnealingLR & SGD & 100 & - & - \\ \hline
\end{tabular}%
}
\label{tab:train-tiered}
\end{table*}

\begin{table*}[thbp]
\centering
\caption{Training configurations and hyperparameters for training on the CUB-200 dataset. ``-'' denotes not applicable.}
\resizebox{\linewidth}{!}{%
\begin{tabular}{ccclcccc}
\hline
Method & Mode & Learning rate & LR scheduler & Optimizer & Epochs & Training episodes & Episode size \\ \hline
\multirow{2}{*}{ANIL} & T (5w1s) & 0.001 & - & Adam & 100 & 2000 & 4 \\
 & T (5w5s) & 0.001 & - & Adam & 100 & 2000 & 4 \\ \hline
\multirow{2}{*}{LEO} & T (5w1s) & 0.0005 & CosineAnnealingLR & Adam & 100 & 2000 & 4 \\
 & T (5w5s) & 0.001 & CosineAnnealingLR & Adam & 100 & 2000 & 1 \\ \hline
\multirow{2}{*}{BOIL} & T (5w1s) & 0.0006 & - & Adam & 100 & 2000 & 4 \\
 & T (5w5s) & 0.0006 & - & Adam & 100 & 2000 & 4 \\ \hline
\multirow{2}{*}{ProtoNet} & T (5w1s) & 0.001 & StepLR(20, 0.5) & Adam & 100 & 2000 & 1 \\
 & T (5w5s) & 0.001 & StepLR(20, 0.5) & Adam & 100 & 2000 & 1 \\ \hline
\multirow{2}{*}{DN4} & T (5w1s) & 0.001 & StepLR(50, 0.5) & Adam & 100 & 2000 & 1 \\
 & T (5w5s) & 0.001 & StepLR(50, 0.5) & Adam & 100 & 2000 & 1 \\ \hline
\multirow{2}{*}{R2D2} & T (5w1s) & 0.1 & CosineAnnealingLR & SGD & 100 & 2000 & 4 \\
 & T (5w5s) & 0.1 & CosineAnnealingLR & SGD & 100 & 2000 & 4 \\ \hline
\multirow{2}{*}{CAN} & T (5w1s) & 0.01 & - & Adam & 100 & 100 & 4 \\
 & T (5w5s) & 0.01 & - & Adam & 100 & 100 & 4 \\ \hline
\multirow{2}{*}{RENet} & T (5w1s) & 0.1 & CosineAnnealingLR & SGD & 100 & 300 & 1 \\
 & T (5w5s) & 0.1 & CosineAnnealingLR & SGD & 100 & 600 & 1 \\ \hline
{Baseline++} & B (128) & 0.01 & CosineAnnealingLR & SGD & 100 & - & - \\ \hline
{RFS} & B (64) & 0.05 & MultiStepLR([60, 80], 0.1) & SGD & 100 & - & - \\ \hline
\end{tabular}%
}
\label{tab:train-cub}
\end{table*}
\section{Experimental Settings}\label{sec:exp-setting-details}
We conducted experiments using three datasets: miniImageNet, tieredImageNet, and CUB-200. Each of these datasets  has training ($\mathcal{D}_{train}$), validation ($\mathcal{D}_{val}$), and test ($\mathcal{D}_{test}$) sets. Numbers of classes and samples in the three sets of the three datasets are shown in \cref{tab:dataset}. 

We trained eight meta-learning based FSC methods with the ResNet-12 backbone using 5-way 1-shot or  5-way 5-shot tasks from each $\mathcal{D}_{train}$ of the three datasets, resulting in a total of 48 models. For the two transfer learning based methods, RFS and Baseline++, we trained them on each $\mathcal{D}_{train}$ of the three datasets using mini-batch stochastic gradient descent. As a result, we trained a total of 54 models. 

To facilitate reproducibility and further research, the training configurations and hyperparameters are provided in \cref{tab:train-mini,tab:train-tiered,tab:train-cub} for training on the miniImageNet, tieredImageNet, and CUB-200 datasets, respectively. We closely followed the settings in \cite{li2021LibFewShot} to train these models. In the ``Mode'' column of these tables, ``T(5w1s)'' denotes that we trained the corresponding model using 5-way 1-shot tasks, ``T(5w5s)'' denotes that we trained the corresponding model using 5-way 5-shot tasks, and ``B (128)'' denotes that we trained the corresponding model using mini-batch stochastic gradient descent with a batch size of 128. In the ``LR scheduler'' column, ``CosineAnnealingLR'' denotes a cosine annealing learning rate scheduler, ``StepLR(20, 0.5)'' denotes a learning rate scheduler which decreases the learning rate after every 20 epochs by multiplying it with 0.5, and ``MultiStepLR([60, 80], 0.1)''  denotes a learning rate scheduler which decreases the learning rate after 60 epochs and 80 epochs by multiplying it with 0.1 each time. The ``Training episodes'' column in these tables denotes the number of tasks used in each epoch. The ``Episode size'' column of these tables denotes the number of tasks jointly used to do a model update.

\begin{table*}[t]
\centering
\caption{Comparison between different techniques used by FewSTAB for constructing the support sets in  5-way 5-shot FSC test tasks. Values in the shaded areas are the accuracy gaps defined as wAcc-R minus wAcc-A. Average drop is the average of accuracy gaps over the ten FSC methods. ``-'' denotes not applicable.}
\resizebox{\linewidth}{!}{%
\begin{tabular}{c|cccc|cccc|cccc}
\hline
 & \multicolumn{4}{c|}{miniImageNet} & \multicolumn{4}{c|}{tieredImageNet} & \multicolumn{4}{c}{CUB-200} \\ \cline{2-13} 
 &  & \multicolumn{3}{c|}{wAcc-A/Acc. gap} &  & \multicolumn{3}{c|}{wAcc-A/Acc. gap} &  & \multicolumn{3}{c}{wAcc-A/Acc. gap} \\ \cline{3-5} \cline{7-9} \cline{11-13} 
\multirow{-3}{*}{Method} & \multirow{-2}{*}{wAcc-R} & SC1 & SC2 & SC3 & \multirow{-2}{*}{wAcc-R} & SC1 & SC2 & SC3 & \multirow{-2}{*}{wAcc-R} & SC1 & SC2 & SC3 \\ \hline
 & & 19.69 & 15.64 & 14.83 & & 19.55 & 14.53 & 13.72 & & 38.73 & 32.39 & 31.63 \\
\multirow{-2}{*}{ANIL} & \multirow{-2}{*}{25.37}& \cellcolor[HTML]{EFEFEF} 5.68 & \cellcolor[HTML]{EFEFEF} 9.73 & \cellcolor[HTML]{EFEFEF}\textbf{ 10.54}  & \multirow{-2}{*}{30.60}& \cellcolor[HTML]{EFEFEF} 11.05 & \cellcolor[HTML]{EFEFEF} 16.07 & \cellcolor[HTML]{EFEFEF}\textbf{ 16.88}  & \multirow{-2}{*}{45.47}& \cellcolor[HTML]{EFEFEF} 6.74 & \cellcolor[HTML]{EFEFEF} 13.08 & \cellcolor[HTML]{EFEFEF}\textbf{ 13.84} \\ \hline
 & & 34.33 & 28.02 & 26.31 & & 40.28 & 30.23 & 29.49 & & 48.04 & 43.07 & 46.62 \\
\multirow{-2}{*}{LEO} & \multirow{-2}{*}{41.33}& \cellcolor[HTML]{EFEFEF} 7.00 & \cellcolor[HTML]{EFEFEF} 13.31 & \cellcolor[HTML]{EFEFEF}\textbf{ 15.02}  & \multirow{-2}{*}{57.22}& \cellcolor[HTML]{EFEFEF} 16.94 & \cellcolor[HTML]{EFEFEF} 26.99 & \cellcolor[HTML]{EFEFEF}\textbf{ 27.73}  & \multirow{-2}{*}{59.76}& \cellcolor[HTML]{EFEFEF} 11.72 & \cellcolor[HTML]{EFEFEF}\textbf{ 16.69} & \cellcolor[HTML]{EFEFEF} 13.14 \\ \hline
 & & 13.88 & 13.46 & 13.09 & & 15.82 & 15.11 & 14.90 & & 18.42 & 17.84 & 19.17 \\
\multirow{-2}{*}{BOIL} & \multirow{-2}{*}{15.21}& \cellcolor[HTML]{EFEFEF} 1.33 & \cellcolor[HTML]{EFEFEF} 1.75 & \cellcolor[HTML]{EFEFEF}\textbf{ 2.12}  & \multirow{-2}{*}{18.55}& \cellcolor[HTML]{EFEFEF} 2.73 & \cellcolor[HTML]{EFEFEF} 3.44 & \cellcolor[HTML]{EFEFEF}\textbf{ 3.65}  & \multirow{-2}{*}{21.33}& \cellcolor[HTML]{EFEFEF} 2.91 & \cellcolor[HTML]{EFEFEF}\textbf{ 3.49} & \cellcolor[HTML]{EFEFEF} 2.16 \\ \hline
 & & 43.37 & 33.40 & 32.07 & & 44.23 & 31.61 & 30.95 & & 67.12 & 59.72 & 60.06 \\
\multirow{-2}{*}{ProtoNet} & \multirow{-2}{*}{51.95}& \cellcolor[HTML]{EFEFEF} 8.58 & \cellcolor[HTML]{EFEFEF} 18.55 & \cellcolor[HTML]{EFEFEF}\textbf{ 19.88}  & \multirow{-2}{*}{62.53}& \cellcolor[HTML]{EFEFEF} 18.30 & \cellcolor[HTML]{EFEFEF} 30.92 & \cellcolor[HTML]{EFEFEF}\textbf{ 31.58}  & \multirow{-2}{*}{75.68}& \cellcolor[HTML]{EFEFEF} 8.56 & \cellcolor[HTML]{EFEFEF}\textbf{ 15.96} & \cellcolor[HTML]{EFEFEF} 15.62 \\ \hline
 & & 36.74 & 28.62 & 27.60 & & 24.32 & 16.50 & 16.07 & & 66.07 & 58.32 & 59.25 \\
\multirow{-2}{*}{DN4} & \multirow{-2}{*}{42.68}& \cellcolor[HTML]{EFEFEF} 5.94 & \cellcolor[HTML]{EFEFEF} 14.06 & \cellcolor[HTML]{EFEFEF}\textbf{ 15.08}  & \multirow{-2}{*}{40.63}& \cellcolor[HTML]{EFEFEF} 16.31 & \cellcolor[HTML]{EFEFEF} 24.13 & \cellcolor[HTML]{EFEFEF}\textbf{ 24.56}  & \multirow{-2}{*}{73.58}& \cellcolor[HTML]{EFEFEF} 7.51 & \cellcolor[HTML]{EFEFEF}\textbf{ 15.26} & \cellcolor[HTML]{EFEFEF} 14.33 \\ \hline
 & & 44.01 & 36.47 & 35.37 & & 43.34 & 31.79 & 31.12 & & 65.12 & 56.88 & 58.66 \\
\multirow{-2}{*}{R2D2} & \multirow{-2}{*}{50.84}& \cellcolor[HTML]{EFEFEF} 6.83 & \cellcolor[HTML]{EFEFEF} 14.37 & \cellcolor[HTML]{EFEFEF}\textbf{ 15.47}  & \multirow{-2}{*}{61.08}& \cellcolor[HTML]{EFEFEF} 17.74 & \cellcolor[HTML]{EFEFEF} 29.29 & \cellcolor[HTML]{EFEFEF}\textbf{ 29.96}  & \multirow{-2}{*}{75.20}& \cellcolor[HTML]{EFEFEF} 10.08 & \cellcolor[HTML]{EFEFEF}\textbf{ 18.32} & \cellcolor[HTML]{EFEFEF} 16.54 \\ \hline
 & & 46.66 & 37.82 & 36.44 & & 45.53 & 32.23 & 31.17 & & 53.91 & 44.91 & 41.31 \\
\multirow{-2}{*}{CAN} & \multirow{-2}{*}{54.23}& \cellcolor[HTML]{EFEFEF} 7.57 & \cellcolor[HTML]{EFEFEF} 16.41 & \cellcolor[HTML]{EFEFEF}\textbf{ 17.79}  & \multirow{-2}{*}{64.19}& \cellcolor[HTML]{EFEFEF} 18.66 & \cellcolor[HTML]{EFEFEF} 31.96 & \cellcolor[HTML]{EFEFEF}\textbf{ 33.02}  & \multirow{-2}{*}{61.61}& \cellcolor[HTML]{EFEFEF} 7.70 & \cellcolor[HTML]{EFEFEF} 16.70 & \cellcolor[HTML]{EFEFEF}\textbf{ 20.30} \\ \hline
 & & 47.48 & 37.60 & 36.19 & & 44.23 & 31.04 & 30.27 & & 63.03 & 53.27 & 52.93 \\
\multirow{-2}{*}{RENet} & \multirow{-2}{*}{56.52}& \cellcolor[HTML]{EFEFEF} 9.04 & \cellcolor[HTML]{EFEFEF} 18.92 & \cellcolor[HTML]{EFEFEF}\textbf{ 20.33}  & \multirow{-2}{*}{63.49}& \cellcolor[HTML]{EFEFEF} 19.26 & \cellcolor[HTML]{EFEFEF} 32.45 & \cellcolor[HTML]{EFEFEF}\textbf{ 33.22}  & \multirow{-2}{*}{71.82}& \cellcolor[HTML]{EFEFEF} 8.79 & \cellcolor[HTML]{EFEFEF} 18.55 & \cellcolor[HTML]{EFEFEF}\textbf{ 18.89} \\ \hline
 & & 37.70 & 30.47 & 29.52 & & 40.95 & 30.74 & 30.01 & & 24.21 & 19.55 & 16.86 \\
\multirow{-2}{*}{Baseline++} & \multirow{-2}{*}{44.94}& \cellcolor[HTML]{EFEFEF} 7.24 & \cellcolor[HTML]{EFEFEF} 14.47 & \cellcolor[HTML]{EFEFEF}\textbf{ 15.42}  & \multirow{-2}{*}{59.06}& \cellcolor[HTML]{EFEFEF} 18.11 & \cellcolor[HTML]{EFEFEF} 28.32 & \cellcolor[HTML]{EFEFEF}\textbf{ 29.05}  & \multirow{-2}{*}{29.84}& \cellcolor[HTML]{EFEFEF} 5.63 & \cellcolor[HTML]{EFEFEF} 10.29 & \cellcolor[HTML]{EFEFEF}\textbf{ 12.98} \\ \hline
 & & 48.17 & 38.33 & 36.85 & & 44.35 & 31.94 & 31.15 & & 64.54 & 54.41 & 54.98 \\
\multirow{-2}{*}{RFS} & \multirow{-2}{*}{55.66}& \cellcolor[HTML]{EFEFEF} 7.49 & \cellcolor[HTML]{EFEFEF} 17.33 & \cellcolor[HTML]{EFEFEF}\textbf{ 18.81}  & \multirow{-2}{*}{62.71}& \cellcolor[HTML]{EFEFEF} 18.36 & \cellcolor[HTML]{EFEFEF} 30.77 & \cellcolor[HTML]{EFEFEF}\textbf{ 31.56}  & \multirow{-2}{*}{74.33}& \cellcolor[HTML]{EFEFEF} 9.79 & \cellcolor[HTML]{EFEFEF}\textbf{ 19.92} & \cellcolor[HTML]{EFEFEF} 19.35 \\ \hline
Average drop  & -  & \cellcolor[HTML]{EFEFEF} 6.67 & \cellcolor[HTML]{EFEFEF} 13.89 & \cellcolor[HTML]{EFEFEF} \textbf{15.05} & -  & \cellcolor[HTML]{EFEFEF} 15.75 & \cellcolor[HTML]{EFEFEF} 25.43 & \cellcolor[HTML]{EFEFEF} \textbf{26.12} & -  & \cellcolor[HTML]{EFEFEF} 7.94 & \cellcolor[HTML]{EFEFEF} \textbf{14.83} & \cellcolor[HTML]{EFEFEF} 14.72\\ \hline
\end{tabular}%
}
\label{tab:ablation-support-construction}
\end{table*}

\begin{table*}[htbp]
\centering
\caption{Comparison between different techniques used by FewSTAB for constructing the query sets in 5-way 5-shot FSC test tasks. Values in the shaded areas are the accuracy gaps defined as wAcc-R minus wAcc-A. Average drop is the average of accuracy gaps over the ten FSC methods. ``-'' denotes not applicable.}
\resizebox{0.95\linewidth}{!}{%
\begin{tabular}{c|cccc|cccc|cccc}
\hline
 & \multicolumn{4}{c|}{miniImageNet} & \multicolumn{4}{c|}{tieredImageNet} & \multicolumn{4}{c}{CUB-200} \\ \cline{2-13} 
 &  & \multicolumn{3}{c|}{wAcc-A/Acc. gap} &  & \multicolumn{3}{c|}{wAcc-A/Acc. gap} &  & \multicolumn{3}{c}{wAcc-A/Acc. gap} \\ \cline{3-5} \cline{7-9} \cline{11-13} 
\multirow{-3}{*}{Method} & \multirow{-2}{*}{wAcc-R} & QC1 & QC2 & QC3 & \multirow{-2}{*}{wAcc-R} & QC1 & QC2 & QC3 & \multirow{-2}{*}{wAcc-R} & QC1 & QC2 & QC3 \\ \hline
 & & 21.61 & 16.37 & 14.83 & & 21.95 & 14.00 & 13.72 & & 39.77 & 32.58 & 31.63 \\
\multirow{-2}{*}{ANIL} & \multirow{-2}{*}{25.37}& \cellcolor[HTML]{EFEFEF} 3.76 & \cellcolor[HTML]{EFEFEF} 9.00 & \cellcolor[HTML]{EFEFEF}\textbf{ 10.54}  & \multirow{-2}{*}{30.60}& \cellcolor[HTML]{EFEFEF} 8.65 & \cellcolor[HTML]{EFEFEF} 16.60 & \cellcolor[HTML]{EFEFEF}\textbf{ 16.88}  & \multirow{-2}{*}{45.47}& \cellcolor[HTML]{EFEFEF} 5.70 & \cellcolor[HTML]{EFEFEF} 12.89 & \cellcolor[HTML]{EFEFEF}\textbf{ 13.84} \\ \hline
 & & 36.04 & 28.36 & 26.31 & & 46.94 & 31.93 & 29.49 & & 56.73 & 48.21 & 46.62 \\
\multirow{-2}{*}{LEO} & \multirow{-2}{*}{41.33}& \cellcolor[HTML]{EFEFEF} 5.29 & \cellcolor[HTML]{EFEFEF} 12.97 & \cellcolor[HTML]{EFEFEF}\textbf{ 15.02}  & \multirow{-2}{*}{57.22}& \cellcolor[HTML]{EFEFEF} 10.28 & \cellcolor[HTML]{EFEFEF} 25.29 & \cellcolor[HTML]{EFEFEF}\textbf{ 27.73}  & \multirow{-2}{*}{59.76}& \cellcolor[HTML]{EFEFEF} 3.03 & \cellcolor[HTML]{EFEFEF} 11.55 & \cellcolor[HTML]{EFEFEF}\textbf{ 13.14} \\ \hline
 & & 14.57 & 13.70 & 13.09 & & 17.61 & 15.37 & 14.90 & & 20.49 & 19.85 & 19.17 \\
\multirow{-2}{*}{BOIL} & \multirow{-2}{*}{15.21}& \cellcolor[HTML]{EFEFEF} 0.64 & \cellcolor[HTML]{EFEFEF} 1.51 & \cellcolor[HTML]{EFEFEF}\textbf{ 2.12}  & \multirow{-2}{*}{18.55}& \cellcolor[HTML]{EFEFEF} 0.94 & \cellcolor[HTML]{EFEFEF} 3.18 & \cellcolor[HTML]{EFEFEF}\textbf{ 3.65}  & \multirow{-2}{*}{21.33}& \cellcolor[HTML]{EFEFEF} 0.84 & \cellcolor[HTML]{EFEFEF} 1.48 & \cellcolor[HTML]{EFEFEF}\textbf{ 2.16} \\ \hline
 & & 44.28 & 34.17 & 32.07 & & 49.58 & 33.56 & 30.95 & & 70.33 & 61.81 & 60.06 \\
\multirow{-2}{*}{ProtoNet} & \multirow{-2}{*}{51.95}& \cellcolor[HTML]{EFEFEF} 7.67 & \cellcolor[HTML]{EFEFEF} 17.78 & \cellcolor[HTML]{EFEFEF}\textbf{ 19.88}  & \multirow{-2}{*}{62.53}& \cellcolor[HTML]{EFEFEF} 12.95 & \cellcolor[HTML]{EFEFEF} 28.97 & \cellcolor[HTML]{EFEFEF}\textbf{ 31.58}  & \multirow{-2}{*}{75.68}& \cellcolor[HTML]{EFEFEF} 5.35 & \cellcolor[HTML]{EFEFEF} 13.87 & \cellcolor[HTML]{EFEFEF}\textbf{ 15.62} \\ \hline
 & & 39.25 & 28.91 & 27.60 & & 28.28 & 17.63 & 16.07 & & 70.37 & 60.61 & 59.25 \\
\multirow{-2}{*}{DN4} & \multirow{-2}{*}{42.68}& \cellcolor[HTML]{EFEFEF} 3.43 & \cellcolor[HTML]{EFEFEF} 13.77 & \cellcolor[HTML]{EFEFEF}\textbf{ 15.08}  & \multirow{-2}{*}{40.63}& \cellcolor[HTML]{EFEFEF} 12.35 & \cellcolor[HTML]{EFEFEF} 23.00 & \cellcolor[HTML]{EFEFEF}\textbf{ 24.56}  & \multirow{-2}{*}{73.58}& \cellcolor[HTML]{EFEFEF} 3.21 & \cellcolor[HTML]{EFEFEF} 12.97 & \cellcolor[HTML]{EFEFEF}\textbf{ 14.33} \\ \hline
 & & 45.68 & 36.99 & 35.37 & & 48.96 & 33.83 & 31.12 & & 69.78 & 60.34 & 58.66 \\
\multirow{-2}{*}{R2D2} & \multirow{-2}{*}{50.84}& \cellcolor[HTML]{EFEFEF} 5.16 & \cellcolor[HTML]{EFEFEF} 13.85 & \cellcolor[HTML]{EFEFEF}\textbf{ 15.47}  & \multirow{-2}{*}{61.08}& \cellcolor[HTML]{EFEFEF} 12.12 & \cellcolor[HTML]{EFEFEF} 27.25 & \cellcolor[HTML]{EFEFEF}\textbf{ 29.96}  & \multirow{-2}{*}{75.20}& \cellcolor[HTML]{EFEFEF} 5.42 & \cellcolor[HTML]{EFEFEF} 14.86 & \cellcolor[HTML]{EFEFEF}\textbf{ 16.54} \\ \hline
 & & 47.83 & 38.16 & 36.44 & & 50.58 & 33.63 & 31.17 & & 54.32 & 42.88 & 41.31 \\
\multirow{-2}{*}{CAN} & \multirow{-2}{*}{54.23}& \cellcolor[HTML]{EFEFEF} 6.40 & \cellcolor[HTML]{EFEFEF} 16.07 & \cellcolor[HTML]{EFEFEF}\textbf{ 17.79}  & \multirow{-2}{*}{64.19}& \cellcolor[HTML]{EFEFEF} 13.61 & \cellcolor[HTML]{EFEFEF} 30.56 & \cellcolor[HTML]{EFEFEF}\textbf{ 33.02}  & \multirow{-2}{*}{61.61}& \cellcolor[HTML]{EFEFEF} 7.29 & \cellcolor[HTML]{EFEFEF} 18.73 & \cellcolor[HTML]{EFEFEF}\textbf{ 20.30} \\ \hline
 & & 49.80 & 38.31 & 36.19 & & 49.86 & 32.76 & 30.27 & & 64.26 & 54.48 & 52.93 \\
\multirow{-2}{*}{RENet} & \multirow{-2}{*}{56.52}& \cellcolor[HTML]{EFEFEF} 6.72 & \cellcolor[HTML]{EFEFEF} 18.21 & \cellcolor[HTML]{EFEFEF}\textbf{ 20.33}  & \multirow{-2}{*}{63.49}& \cellcolor[HTML]{EFEFEF} 13.63 & \cellcolor[HTML]{EFEFEF} 30.73 & \cellcolor[HTML]{EFEFEF}\textbf{ 33.22}  & \multirow{-2}{*}{71.82}& \cellcolor[HTML]{EFEFEF} 7.56 & \cellcolor[HTML]{EFEFEF} 17.34 & \cellcolor[HTML]{EFEFEF}\textbf{ 18.89} \\ \hline
 & & 39.51 & 31.26 & 29.52 & & 47.08 & 31.86 & 30.01 & & 27.47 & 18.62 & 16.86 \\
\multirow{-2}{*}{Baseline++} & \multirow{-2}{*}{44.94}& \cellcolor[HTML]{EFEFEF} 5.43 & \cellcolor[HTML]{EFEFEF} 13.68 & \cellcolor[HTML]{EFEFEF}\textbf{ 15.42}  & \multirow{-2}{*}{59.06}& \cellcolor[HTML]{EFEFEF} 11.98 & \cellcolor[HTML]{EFEFEF} 27.20 & \cellcolor[HTML]{EFEFEF}\textbf{ 29.05}  & \multirow{-2}{*}{29.84}& \cellcolor[HTML]{EFEFEF} 2.37 & \cellcolor[HTML]{EFEFEF} 11.22 & \cellcolor[HTML]{EFEFEF}\textbf{ 12.98} \\ \hline
 & & 48.87 & 39.48 & 36.85 & & 49.99 & 33.64 & 31.15 & & 67.60 & 56.60 & 54.98 \\
\multirow{-2}{*}{RFS} & \multirow{-2}{*}{55.66}& \cellcolor[HTML]{EFEFEF} 6.79 & \cellcolor[HTML]{EFEFEF} 16.18 & \cellcolor[HTML]{EFEFEF}\textbf{ 18.81}  & \multirow{-2}{*}{62.71}& \cellcolor[HTML]{EFEFEF} 12.72 & \cellcolor[HTML]{EFEFEF} 29.07 & \cellcolor[HTML]{EFEFEF}\textbf{ 31.56}  & \multirow{-2}{*}{74.33}& \cellcolor[HTML]{EFEFEF} 6.73 & \cellcolor[HTML]{EFEFEF} 17.73 & \cellcolor[HTML]{EFEFEF}\textbf{ 19.35} \\ \hline
Average drop  & -  & \cellcolor[HTML]{EFEFEF} 5.13 & \cellcolor[HTML]{EFEFEF} 13.30 & \cellcolor[HTML]{EFEFEF} \textbf{15.05} & -  & \cellcolor[HTML]{EFEFEF} 10.92 & \cellcolor[HTML]{EFEFEF} 24.19 & \cellcolor[HTML]{EFEFEF} \textbf{26.12} & -  & \cellcolor[HTML]{EFEFEF} 4.75 & \cellcolor[HTML]{EFEFEF} 13.27 & \cellcolor[HTML]{EFEFEF} \textbf{14.72}\\ \hline
\end{tabular}%
}
\label{tab:ablation-query-construction}
\end{table*}

% \begin{table*}[htbp]
% \centering
% %\resizebox{\columnwidth}{!}{%
% \begin{tabular}{lccc}
% \hline
% \multirow{2}{*}{Random selection from} & \multicolumn{3}{c}{Avg. drop (\%)} \\ \cline{2-4} 
%  & miniImageNet & tieredImageNet & CUB-200 \\ \hline
% All samples & 6.34 &  14.62& 8.04 \\
% Samples with targeted spurious attributes & 12.96 & 23.33 & 14.48 \\
% Samples with mutually exclusive targeted spurious attributes & 15.05 & 26.12 & 14.72 \\ \hline
% \end{tabular}%
% %}
% \caption{Comparison between different techniques used by FewSTAB for constructing the support sets in FSC tasks. Average drop is the average of wAcc-R minus wAcc-A obtained on 5-way 5-shot tasks over the ten FSC methods.}
% \label{tab:fsc-support-task-construction-methods-all}
% \end{table*}

% \begin{table*}[htbp]
% \centering
% %\resizebox{\linewidth}{!}{%
% \begin{tabular}{cccccc}
% \hline
% \multicolumn{2}{c}{attribute-based sample selection} & \multirow{2}{*}{Query sample selection} & \multicolumn{3}{c}{Avg. drop (\%)} \\ \cline{1-2}\cline{4-6}
% Intra-class & Inter-class &  &  miniImageNet & tieredImageNet & CUB-200\\ \hline
% $\checkmark$ &  &  & 5.13 &10.92 & 4.75\\
% $\checkmark$ & $\checkmark$ &  & 13.30 & 24.18& 13.27\\
% $\checkmark$ & $\checkmark$ & $\checkmark$ & 15.05& 26.12& 14.72\\ \hline
% \end{tabular}%
% %}
% \caption{Comparison between different techniques used by FewSTAB for constructing the query sets in FSC tasks. Average drop is the average of wAcc-R minus wAcc-A obtained on 5-way 5-shot tasks over the ten FSC methods.}
% \label{tab:fsc-task-construction-methods-all}
% \end{table*}

\section{Additional Experimental Results} \label{sec:additional-ablation-studies}

\subsection{Ablation Studies}
\label{sec:appendix-ablation-studies}
\noindent\textbf{Support set construction methods:}
To construct the support set in an FSC test task, FewSTAB randomly selects samples that have \textit{mutually exclusive} spurious attributes across the randomly selected classes, which is illustrated in \cref{fig:method-overview}(a) and formally described in \cref{sec:spurious-test-task} in the main paper. To further show the effectiveness of this construction method, we keep the techniques for constructing the query set in an FSC test task, and report in \cref{tab:ablation-support-construction} the results of two alternatives for constructing the support set: randomly selecting samples of the selected classes (\textbf{SC1})  and randomly selecting samples with targeted attributes for selected classes with no further constraints on the selected samples (\textbf{SC2}). We also include the results of the proposed one: randomly selecting samples with mutually exclusive targeted attributes across the selected classes (\textbf{SC3}) in \cref{tab:ablation-support-construction}. 
A larger average drop in \cref{tab:ablation-support-construction} indicates that the corresponding support set construction method is more effective in revealing robustness of few-shot classifiers to spurious bias. We observe that the third technique SC3, which is used by FewSTAB, achieves the largest average accuracy drop among the techniques compared on the miniImageNet and tieredImageNet datasets and achieves a comparable drop to SC2 on the CUB-200 dataset due to the limited number of detected attributes in this dataset.

\noindent\textbf{Query set construction methods:}
There are three techniques used by FewSTAB to construct the query set in a task: the intra-class attribute-based sample selection (\textbf{QC1}), the inter-class attribute-based sample selection (\textbf{QC2}), which is a special case of the intra-class attribute-based sample selection, and the query sample selection  (\textbf{QC3}). We have done an ablation study on the effectiveness of the three techniques in \cref{tab:fsc-task-construction-methods} in the main paper using the miniImageNet dataset. Here, we include the results on all the three datasets in \cref{tab:ablation-query-construction}. 
We observe that all the three proposed techniques in FewSTAB are effective with positive accuracy drops for all the ten FSC methods on the three datasets. Moreover, using the inter-class attribute-based sample selection significantly improves the average drops of the intra-class attribute-based sample selection, with 8.17\%, 13.26\%, and 8.52\% absolute gains on the miniImageNet, tieredImageNet, and CUB-200 datasets, respectively.

\subsection{Scatter Plots of wAcc-A versus Acc}
\label{sec:appendix-scatter-plots}
\begin{figure*}[htbp]
    \centering
    \includegraphics[width=\linewidth]{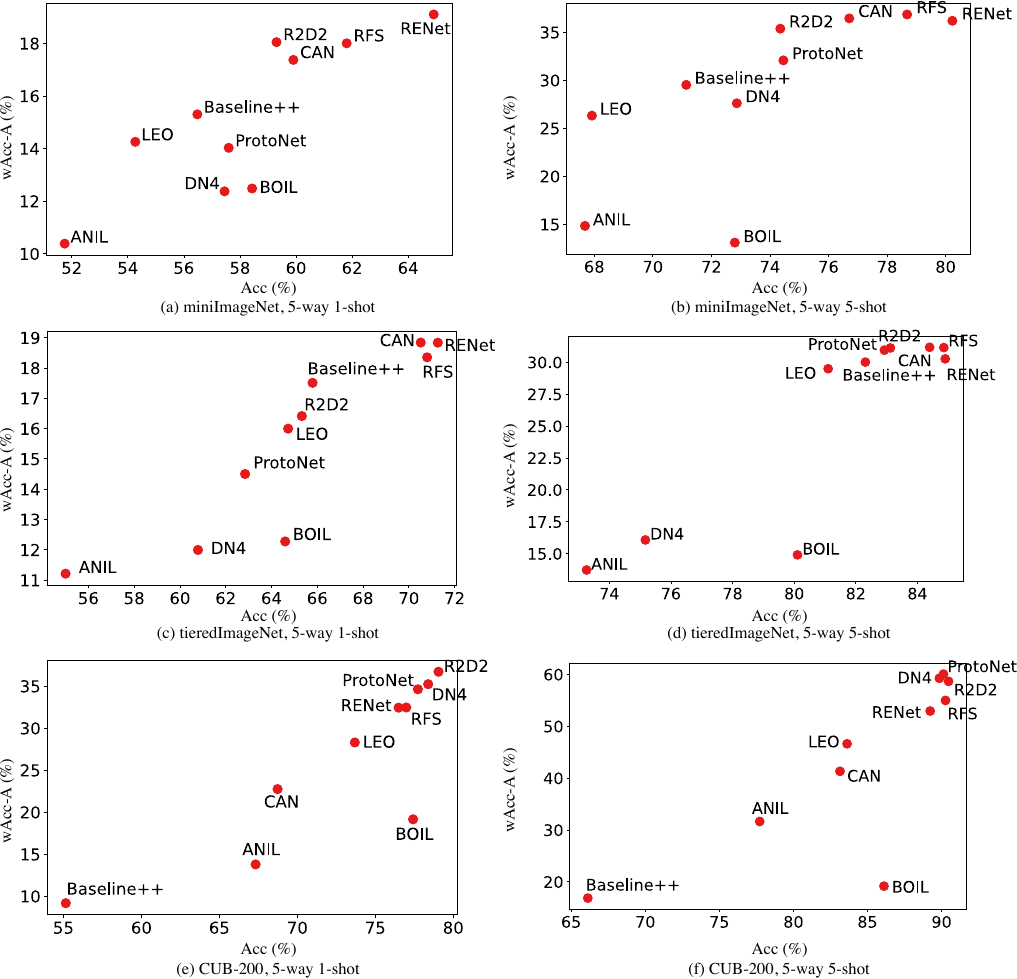}
    \caption{Scatter plots of wAcc-A versus Acc of the ten FSC methods tested with 5-way 1/5-shot FewSTAB and randomly constructed tasks from the miniImageNet, tieredImageNet, and CUB-200 datasets, respectively. All methods are trained and tested with the same shot number.}
    \label{fig:wAcc-A-avg-all}
\end{figure*}

We show the scatter plots of wAcc-A versus Acc (standard accuracy) of the ten FSC methods when they are tested with FewSTAB and randomly constructed FSC test tasks, respectively, on the three datasets in \cref{fig:wAcc-A-avg-all} (exact values are shown in \cref{tab:average-worst-classification}). We observe that an FSC method having a higher Acc does not necessarily have a higher wAcc-A. For example, in  \cref{fig:wAcc-A-avg-all}(a), BOIL  has a higher Acc but a lower wAcc-A than ProtoNet, LEO, and Baseline++. Moreover, we observe that in \cref{fig:wAcc-A-avg-all}(b) and (d), for methods that achieve high standard accuracies, e.g., for the top-5 methods in terms of Acc, their relative increments in wAcc-A are small (with differences smaller than 1\%) compared with their relative increments in Acc. In other words, methods with higher standard accuracies do not necessarily learn more robust decision rules, since their wAcc-A values remain comparable to those with lower Acc values.

The values of Acc and wAcc-A on the fine-grained dataset CUB-200 in \cref{fig:wAcc-A-avg-all}(e) and (f) show a different pattern from those in \cref{fig:wAcc-A-avg-all}(c) and (d). More specifically, methods that achieve high Acc values, e.g., R2D2, ProtoNet, DN4, RENet, and RFS, tend to have comparable relative increments in wAcc-A compared with their relative increments in Acc. This indicates that on a fine-grained dataset, which does not have many spurious attributes, an FSC method with a higher Acc also tends to have a higher wAcc-A or improved robustness to spurious bias. 

In summary, our framework, FewSTAB, reveals new robustness patterns of FSC methods in different evaluation settings.

\renewcommand{\minval}{0}
\renewcommand{\maxval}{100}
\renewcommand{\opacity}{25}
\begin{table*}[t]
\centering
\caption{Standard accuracies (Acc) and class-wise worst accuracies obtained with FewSTAB (wAcc-A) with 95\% confidence intervals of the ten FSC methods on  miniImageNet, tieredImageNet, and CUB datasets. Numbers in the Shot column indicate that the models are both trained (if applicable) and tested on 5-way 1- or 5-shot tasks. Darker colors indicate higher values.}
\resizebox{\linewidth}{!}{%
\begin{tabular}{c|l|c@{\hskip 0.2cm}c@{\hskip 0.2cm}c@{\hskip 0.2cm}c@{\hskip 0.2cm}c@{\hskip 0.2cm}c}
\hline
\multirow{2}{*}{Shot} & \multicolumn{1}{c|}{\multirow{2}{*}{Method}} & \multicolumn{2}{c}{miniImageNet} & \multicolumn{2}{c}{tieredImageNet} & \multicolumn{2}{c}{CUB-200} \\ \cline{3-8} 
&  & Acc & wAcc-A &  Acc & wAcc-A &  Acc & wAcc-A \\ \hline
\multirow{10}{*}{1} & ANIL & \gradient{51.75}$\pm$0.39 & \gradient{10.38}$\pm$0.30 & \gradient{55.00}$\pm$0.45 & \gradient{11.21}$\pm$0.30 & \gradient{67.32}$\pm$0.45 & \gradient{13.78}$\pm$0.40 \\
& LEO & \gradient{54.27}$\pm$0.38 & \gradient{14.26}$\pm$0.46 & \gradient{64.73}$\pm$0.46 & \gradient{16.00}$\pm$0.55 & \gradient{73.68}$\pm$0.42 & \gradient{28.29}$\pm$0.80 \\
& BOIL & \gradient{58.43}$\pm$0.39 & \gradient{12.48}$\pm$0.23 & \gradient{64.60}$\pm$0.43 & \gradient{12.27}$\pm$0.21 & \gradient{77.42}$\pm$0.39 & \gradient{19.15}$\pm$0.29 \\
& ProtoNet & \gradient{57.60}$\pm$0.38 & \gradient{14.03}$\pm$0.49 & \gradient{62.85}$\pm$0.44 & \gradient{14.50}$\pm$0.50 & \gradient{77.73}$\pm$0.39 & \gradient{34.62}$\pm$0.85 \\
& DN4 & \gradient{57.45}$\pm$0.36 & \gradient{12.37}$\pm$0.46 & \gradient{60.79}$\pm$0.42 & \gradient{11.99}$\pm$0.47 & \gradient{78.39}$\pm$0.38 & \gradient{35.22}$\pm$0.86 \\
& R2D2 & \gradient{59.30}$\pm$0.39 & \gradient{18.05}$\pm$0.53 & \gradient{65.33}$\pm$0.44 & \gradient{16.41}$\pm$0.54 & \gradient{79.05}$\pm$0.38 & \gradient{36.70}$\pm$0.90 \\
& CAN & \gradient{59.91}$\pm$0.38 & \gradient{17.37}$\pm$0.53 & \gradient{70.52}$\pm$0.43 & \gradient{18.84}$\pm$0.60 & \gradient{68.73}$\pm$0.41 & \gradient{22.74}$\pm$0.72 \\
& RENet & \gradient{64.91}$\pm$0.38 & \gradient{19.10}$\pm$0.57 & \gradient{71.27}$\pm$0.42 & \gradient{18.83}$\pm$0.61 & \gradient{76.49}$\pm$0.36 & \gradient{32.43}$\pm$0.81 \\
& Baseline++ & \gradient{56.48}$\pm$0.37 & \gradient{15.30}$\pm$0.48 & \gradient{65.79}$\pm$0.42 & \gradient{17.51}$\pm$0.54 & \gradient{55.15}$\pm$0.44 & \gradient{9.17}$\pm$0.47 \\
& RFS & \gradient{61.81}$\pm$0.35 & \gradient{18.00}$\pm$0.53 & \gradient{70.80}$\pm$0.42 & \gradient{18.35}$\pm$0.60 & \gradient{76.99}$\pm$0.35 & \gradient{32.45}$\pm$0.80 \\ \hline
\multirow{10}{*}{5} & ANIL& \gradient{67.68}$\pm$0.33 & \gradient{14.83}$\pm$0.40 & \gradient{73.26}$\pm$0.35 & \gradient{13.72}$\pm$0.39 & \gradient{77.72}$\pm$0.34 & \gradient{31.63}$\pm$0.55 \\
& LEO& \gradient{67.92}$\pm$0.32 & \gradient{26.31}$\pm$0.59 & \gradient{81.10}$\pm$0.34 & \gradient{29.49}$\pm$0.72 & \gradient{83.62}$\pm$0.30 & \gradient{46.62}$\pm$0.82 \\
& BOIL& \gradient{72.80}$\pm$0.29 & \gradient{13.09}$\pm$0.22 & \gradient{80.11}$\pm$0.32 & \gradient{14.90}$\pm$0.22 & \gradient{86.11}$\pm$0.26 & \gradient{19.17}$\pm$0.28 \\
& ProtoNet& \gradient{74.46}$\pm$0.28 & \gradient{32.07}$\pm$0.58 & \gradient{82.93}$\pm$0.31 & \gradient{30.95}$\pm$0.70 & \gradient{90.13}$\pm$0.21 & \gradient{60.06}$\pm$0.74 \\
& DN4& \gradient{72.87}$\pm$0.29 & \gradient{27.60}$\pm$0.58 & \gradient{75.17}$\pm$0.36 & \gradient{16.07}$\pm$0.62 & \gradient{89.85}$\pm$0.21 & \gradient{59.25}$\pm$0.77 \\
& R2D2& \gradient{74.36}$\pm$0.29 & \gradient{35.37}$\pm$0.59 & \gradient{83.12}$\pm$0.30 & \gradient{31.12}$\pm$0.72 & \gradient{90.47}$\pm$0.21 & \gradient{58.66}$\pm$0.82 \\
& CAN& \gradient{76.71}$\pm$0.28 & \gradient{36.44}$\pm$0.65 & \gradient{84.40}$\pm$0.29 & \gradient{31.17}$\pm$0.76 & \gradient{83.14}$\pm$0.27 & \gradient{41.31}$\pm$0.74 \\
& RENet& \gradient{80.23}$\pm$0.26 & \gradient{36.19}$\pm$0.63 & \gradient{84.90}$\pm$0.28 & \gradient{30.27}$\pm$0.76 & \gradient{89.23}$\pm$0.21 & \gradient{52.93}$\pm$0.82 \\
& Baseline++& \gradient{71.14}$\pm$0.30 & \gradient{29.52}$\pm$0.57 & \gradient{82.31}$\pm$0.31 & \gradient{30.01}$\pm$0.72 & \gradient{66.12}$\pm$0.35 & \gradient{16.86}$\pm$0.52 \\
& RFS& \gradient{78.69}$\pm$0.26 & \gradient{36.85}$\pm$0.64 & \gradient{84.86}$\pm$0.29 & \gradient{31.15}$\pm$0.76 & \gradient{90.26}$\pm$0.20 & \gradient{54.98}$\pm$0.81 \\ \bottomrule
\end{tabular}%
}
\label{tab:average-worst-classification}
\end{table*}

\begin{table*}[t]
\centering
\caption{Comparison between wAcc-A calculated over 5-way 1/5-shot tasks obtained using Vit-GPT2 and using BLIP. We calculated wAcc-A for ten FSC methods on  miniImageNet, tieredImageNet, and CUB datasets. Numbers in the Shot column indicate that the models are both trained (if applicable) and tested on 1- or 5-shot tasks. Darker colors indicate higher values.}
\resizebox{\linewidth}{!}{%
\begin{tabular}{c|l|c@{\hskip 0.2cm}c@{\hskip 0.2cm}c@{\hskip 0.2cm}c@{\hskip 0.2cm}c@{\hskip 0.2cm}c}
\hline
\multirow{2}{*}{Shot} & \multicolumn{1}{c|}{\multirow{2}{*}{Method}}  & \multicolumn{2}{c}{miniImageNet} & \multicolumn{2}{c}{tieredImageNet} & \multicolumn{2}{c}{CUB-200} \\ \cline{3-8} 
&  & ViT-GPT2 & BLIP &  ViT-GPT2 &BLIP &  ViT-GPT2 & BLIP \\ \hline
\multirow{10}{*}{1} & ANIL & \gradient{10.38}$\pm$0.30 & \gradient{10.39}$\pm$0.29 & \gradient{11.21}$\pm$0.30 & \gradient{10.76}$\pm$0.29 & \gradient{13.78}$\pm$0.40 & \gradient{14.74}$\pm$0.41 \\
& LEO & \gradient{14.26}$\pm$0.46 & \gradient{14.38}$\pm$0.45 & \gradient{16.00}$\pm$0.55 & \gradient{14.34}$\pm$0.52 & \gradient{28.29}$\pm$0.80 & \gradient{31.06}$\pm$0.80 \\
& BOIL & \gradient{12.48}$\pm$0.23 & \gradient{12.51}$\pm$0.22 & \gradient{12.27}$\pm$0.21 & \gradient{11.65}$\pm$0.21 & \gradient{19.15}$\pm$0.29 & \gradient{20.35}$\pm$0.29 \\
& ProtoNet & \gradient{14.03}$\pm$0.49 & \gradient{13.50}$\pm$0.46 & \gradient{14.50}$\pm$0.50 & \gradient{13.25}$\pm$0.50 & \gradient{34.62}$\pm$0.85 & \gradient{38.63}$\pm$0.81 \\
& DN4 & \gradient{12.37}$\pm$0.46 & \gradient{12.86}$\pm$0.46 & \gradient{11.99}$\pm$0.47 & \gradient{11.21}$\pm$0.46 & \gradient{35.22}$\pm$0.86 & \gradient{39.51}$\pm$0.82 \\
& R2D2 & \gradient{18.05}$\pm$0.53 & \gradient{17.66}$\pm$0.51 & \gradient{16.41}$\pm$0.54 & \gradient{15.01}$\pm$0.53 & \gradient{36.70}$\pm$0.90 & \gradient{40.61}$\pm$0.84 \\
& CAN & \gradient{17.37}$\pm$0.53 & \gradient{16.89}$\pm$0.51 & \gradient{18.84}$\pm$0.60 & \gradient{17.43}$\pm$0.61 & \gradient{22.74}$\pm$0.72 & \gradient{24.23}$\pm$0.71 \\
& RENet & \gradient{19.10}$\pm$0.57 & \gradient{18.80}$\pm$0.54 & \gradient{18.83}$\pm$0.61 & \gradient{17.29}$\pm$0.60 & \gradient{32.43}$\pm$0.81 & \gradient{36.12}$\pm$0.82 \\
& Baseline++ & \gradient{15.30}$\pm$0.48 & \gradient{15.06}$\pm$0.46 & \gradient{17.51}$\pm$0.54 & \gradient{15.60}$\pm$0.52 & \gradient{9.17}$\pm$0.47 & \gradient{10.42}$\pm$0.50 \\
& RFS & \gradient{18.00}$\pm$0.53 & \gradient{17.43}$\pm$0.50 & \gradient{18.35}$\pm$0.60 & \gradient{16.81}$\pm$0.57 & \gradient{32.45}$\pm$0.80 & \gradient{35.43}$\pm$0.79 \\ \hline
\multirow{10}{*}{5} & ANIL& \gradient{14.83}$\pm$0.40 & \gradient{13.67}$\pm$0.38 & \gradient{13.72}$\pm$0.39 & \gradient{12.57}$\pm$0.37 & \gradient{31.63}$\pm$0.55 & \gradient{33.01}$\pm$0.56 \\
& LEO& \gradient{26.31}$\pm$0.59 & \gradient{24.79}$\pm$0.57 & \gradient{29.49}$\pm$0.72 & \gradient{27.92}$\pm$0.70 & \gradient{46.62}$\pm$0.82 & \gradient{49.97}$\pm$0.81 \\
& BOIL& \gradient{13.09}$\pm$0.22 & \gradient{12.79}$\pm$0.22 & \gradient{14.90}$\pm$0.22 & \gradient{14.63}$\pm$0.22 & \gradient{19.17}$\pm$0.28 & \gradient{20.03}$\pm$0.27 \\
& ProtoNet& \gradient{32.07}$\pm$0.58 & \gradient{29.28}$\pm$0.57 & \gradient{30.95}$\pm$0.70 & \gradient{28.51}$\pm$0.68 & \gradient{60.06}$\pm$0.74 & \gradient{64.67}$\pm$0.64 \\
& DN4& \gradient{27.60}$\pm$0.58 & \gradient{25.28}$\pm$0.57 & \gradient{16.07}$\pm$0.62 & \gradient{14.98}$\pm$0.58 & \gradient{59.25}$\pm$0.77 & \gradient{65.61}$\pm$0.67 \\
& R2D2& \gradient{35.37}$\pm$0.59 & \gradient{31.81}$\pm$0.59 & \gradient{31.12}$\pm$0.72 & \gradient{29.50}$\pm$0.68 & \gradient{58.66}$\pm$0.82 & \gradient{64.02}$\pm$0.77 \\
& CAN& \gradient{36.44}$\pm$0.65 & \gradient{33.81}$\pm$0.62 & \gradient{31.17}$\pm$0.76 & \gradient{29.28}$\pm$0.72 & \gradient{41.31}$\pm$0.74 & \gradient{43.10}$\pm$0.73 \\
& RENet& \gradient{36.19}$\pm$0.63 & \gradient{33.76}$\pm$0.63 & \gradient{30.27}$\pm$0.76 & \gradient{28.71}$\pm$0.72 & \gradient{52.93}$\pm$0.82 & \gradient{60.29}$\pm$0.74 \\
& Baseline++& \gradient{29.52}$\pm$0.57 & \gradient{27.17}$\pm$0.55 & \gradient{30.01}$\pm$0.72 & \gradient{28.20}$\pm$0.70 & \gradient{16.86}$\pm$0.52 & \gradient{17.25}$\pm$0.53 \\
& RFS& \gradient{36.85}$\pm$0.64 & \gradient{34.72}$\pm$0.62 & \gradient{31.15}$\pm$0.76 & \gradient{29.29}$\pm$0.72 & \gradient{54.98}$\pm$0.81 & \gradient{62.33}$\pm$0.69 \\ \bottomrule
\end{tabular}%
}
\label{tab:worst-classification-vlm-comparison}
\end{table*}

\begin{table}[t]
\centering
\caption{Results on the miniImageNet dataset. V: ViT-GPT2, B: BLIP. All input images are resized to 84$\times$84.}
%\resizebox{\linewidth}{!}{%
\begin{tabular}{lcccc}
\hline
Method  & Shot & wAcc-R     & wAcc-A (V) & wAcc-A (B) \\ \hline
UniSiam & 1    & $21.26_{\pm0.48}$ & $13.52_{\pm0.43}$ & $13.49_{\pm0.42}$ \\
PsCo   & 1    & $21.50_{\pm0.47}$ & $14.30_{\pm0.40}$ & $12.46_{\pm0.37}$ \\
BECLR  & 1    & $35.57_{\pm0.80}$ & $23.60_{\pm0.83}$ & $22.42_{\pm0.82}$ \\ \hline
UniSiam & 5    & $45.60_{\pm0.52}$ & $27.76_{\pm0.57}$ & $25.42_{\pm0.56}$ \\
PsCo  & 5    & $42.15_{\pm0.52}$ & $25.54_{\pm0.52}$ & $22.64_{\pm0.49}$ \\
BECLR & 5    & $55.20_{\pm0.49}$ & $37.32_{\pm0.66}$ & $33.42_{\pm0.68}$ \\ \hline
\end{tabular}%
%}
\label{tab:new-results}
\end{table}

\subsection{Effectiveness of FewSTAB: More Results}
\label{sec:effectiveness-more-results}
\noindent\textbf{Results on more recent methods.}
Note that our method selection in \cref{tab:worst-classification} aims to cover \textit{diverse} methods and allow for \textit{rigorous} comparison in the \textit{same} setting. Importantly, our method is general and can continue to evaluate emerging methods. To demonstrate, we provide results on recent methods, namely UniSiam \cite{lu2022self}, PsCo \cite{jangunsupervised}, and BECLR \cite{poulakakisbeclr}. FewSTAB uncovers that, even the state-of-the-art methods still suffer from spurious bias as we observe large gaps between wAcc-R and wAcc-A (Table \ref{tab:new-results}),  when we explicitly construct the test tasks to have spurious correlations.  This also shows that FewSTAB is effective for various FSC methods.

\noindent\textbf{Results on IFSL.}  Interventional few-shot learning (IFSL) \cite{yue2020interventional} is a method that specifically addresses spurious correlations in few-shot classification. We follow the settings in \cite{yue2020interventional} and report the results of MAML \cite{finn2017model}, MN \cite{Vinyals2016Matching}, SIB \cite{huempirical}, and MTL \cite{sun2019meta} in Table \ref{tab:results-ifsl}, where ``Base'' refers to one of the four methods, ``+IFSL'' denotes using IFSL on top of ``Base'', and the better performance between the two is in bold. Overall, IFSL is effective in mitigating spurious bias in few-shot classifiers except for some methods, e.g. SIB. This shows that FewSTAB can reveal the improvement made to mitigate spurious bias.

\begin{table}[t]
\centering
\caption{wAcc-A comparison (\%) on the miniImageNet dataset.}
%\resizebox{0.92\linewidth}{!}{%
\begin{tabular}{ccccc}
\hline
\multirow{2}{*}{Method} & \multicolumn{2}{c}{1-shot}                      & \multicolumn{2}{c}{5-shot}        \\ \cline{2-5} 
                        & Base                          & +IFSL           & Base            & +IFSL           \\ \hline
MAML                    &            \textbf{13.29}$_{\pm0.55}$   &      $12.05_{\pm 0.56}$           &         $28.70_{\pm0.69}$        &     \textbf{29.82}$_{\pm0.76}$            \\
MN                      & $17.40_{\pm0.62}$ &    \textbf{17.72}$_{\pm 0.63} $     & $30.48_{\pm0.73}$  &     \textbf{31.51}$_{\pm 0.75}$           \\
SIB                     & \textbf{30.09}$_{\pm1.04}$                & $27.10_{\pm 0.97}$ & \textbf{46.73}$_{\pm 0.96}$ & $46.66_{\pm 0.95}$ \\
MTL                     & $37.29_{\pm 0.57}$            &      \textbf{40.22}$_{\pm 0.57}$            & $49.49_{\pm0.58}$  &         \textbf{52.66}$_{\pm 0.58}$        \\ \hline
\end{tabular}
%}
\label{tab:results-ifsl}
\end{table}

\subsection{Robustness of FewSTAB with Different VLMs}
\label{sec:appendix-different-vlms}
We instantiated our FewSTAB with a pre-trained ViT-GPT2 and a pre-trained BLIP, respectively. We calculated the wAcc-A on FSC test tasks constructed by FewSTAB with the two VLMs on the miniImageNet, tieredImageNet, and CUB-200 datasets, respectively.

\noindent\textbf{Effects on individual and relative measurements.} We observe from \cref{tab:worst-classification-vlm-comparison} that FewSTAB with BLIP produces lower wAcc-A than with ViT-GPT2 on the miniImageNet and tieredImageNet datasets. This indicates that FewSTAB with BLIP is more effective in uncovering the robustness of few-shot classifiers to spurious bias. We reason that BLIP can identify more attributes than ViT-GPT2 (\cref{tab:feat-statistics}) and therefore more spurious correlations can be formulated by our FewSTAB. 
However, on the fine-grained CUB-200 dataset, which contains different bird classes, FewSTAB with BLIP is less effective than with ViT-GPT2. Although BLIP can identify more attributes than ViT-GPT2 in this fine-grained dataset, it may also detect more attributes related to classes.  To validate this, we first found a set of attributes $\mathcal{U}_{\text{BLIP}}$ unique to BLIP from all the attributes  $\mathcal{A}_{\text{BLIP}}$ detected by BLIP, and a set of attributes $\mathcal{U}_{\text{ViT-GPT2}}$ unique to ViT-GPT2 from all the attributes $\mathcal{A}_{\text{ViT-GPT2}}$ detected by ViT-GPT2. Specifically, we have $\mathcal{U}_{\text{BLIP}}=\mathcal{A}_{\text{BLIP}}-\mathcal{A}_{\text{ViT-GPT2}}$, and $\mathcal{U}_{\text{ViT-GPT2}}=\mathcal{A}_{\text{ViT-GPT2}}-\mathcal{A}_{\text{BLIP}}$. Then, we found in $\mathcal{U}_{\text{BLIP}}$ and $\mathcal{U}_{\text{ViT-GPT2}}$ how many attributes contain ``bird'', ``beak'', ``wing'', ``breast'', ``tail'', or ``mouth'', which are all related to the concept of a bird. We found that there are 11 attributes, or 8.5\% of total attributes in $\mathcal{U}_{\text{BLIP}}$ that are related to a bird. While there is only 1 attribute (2.4\% of total attributes) in $\mathcal{U}_{\text{ViT-GPT2}}$ that is related to a bird. Due to the limited capability of BLIP, these class-related attributes cannot be detected in all the images. Hence, although these attributes are not spurious, they are treated as spurious attributes and used by FewSTAB to construct FSC test tasks. In this case, FewSTAB becomes ineffective in revealing the spurious bias in few-shot classifiers since the classifiers can exploit spurious correlations in the tasks to achieve high accuracies.
Nevertheless, from the perspective of comparing the robustness of different FSC methods to spurious bias, the test tasks constructed by FewSTAB using different VLMs can reveal consistent ranks in terms of wAcc-A for different FSC methods (\cref{tab:spearman-corr-vlm}).

\noindent\textbf{Detection accuracies of VLMs.} Using different VLMs may generate different sets of attributes. Some sets of attributes may not exactly reflect the data being described, resulting in low detection accuracies. For example, some attributes are not identified by a VLM or the identified attributes do not match with the ground truth attributes. To analyze how the detection accuracy of a VLM affects our framework, we show  in \cref{tab:detection-accuracy-vlms} the detection accuracies of the two VLMs that we used in our paper along with the Spearman's rank correlation coefficients between the evaluation results on the ten FSC methods based on the two VLMs. To calculate the detection accuracy of a VLM without the labor-intensive human labeling, we use the outputs of another VLM as the ground truth. Specifically, for the $i$'th image, we have two detected sets of attributes, $\mathcal{A}_{query}^i$ and $\mathcal{A}_{ref}^i$, representing the attributes from a VLM being evaluated and the ones from another VLM serving as the ground truth attributes. The detection accuracy  is calculated as follows:
\begin{align}
    Acc(VLM_{query},VLM_{ref})=\frac{1}{|\mathcal{D}_{test}|}\sum_{i=1}^{N_{test}}\frac{|\mathcal{A}_{query}^i\cap \mathcal{A}_{ref}^i|}{|\mathcal{A}_{ref}^i|},
\end{align}
where $N_{test}=|\mathcal{D}_{test}|$,  and $|\cdot|$ denotes the size of a set. For example, to calculate the detection accuracy of ViT-GPT2, we set $VLM_{query}=$ViT-GPT2 and $VLM_{ref}=$BLIP. From \cref{tab:detection-accuracy-vlms}, we observe that the detection accuracies of the two VLMs are not high, indicating that the attributes identified by the two VLMs are very different. However, the two VLMs are well-established in practice and can identify many attributes from images (\cref{tab:feat-statistics}). The correlation coefficients in \cref{tab:detection-accuracy-vlms} indicate that for well-established VLMs, the detection accuracies have little impact on the comparison of robustness to spurious bias between different FSC methods.

\begin{table}[t]
\centering
\caption{Detection accuracies of the ViT-GPT2 and BLIP along with the Spearman's rank correlation coefficients between the results based on the two VLMs.}
%\resizebox{\linewidth}{!}{%
\begin{tabular}{ccccc}
\hline
\multirow{2}{*}{VLM} & \multicolumn{2}{c}{Detection accuracy} & \multicolumn{2}{c}{Spearman's rank correlation coefficient} \\ \cline{2-5} 
                     & ViT-GPT2            & BLIP             & 1-shot                     & 5-shot                    \\ \hline
miniImageNet         & 34.46               & 31.42            & 0.98                       & 1.0                       \\
tieredImageNet       & 35.04               & 32.00            & 1.0                        & 0.99                      \\
CUB-200              & 70.12               & 59.28            & 1.0                        & 0.98                      \\ \hline
\end{tabular}%
%}
\label{tab:detection-accuracy-vlms}
\end{table}

\section{Tasks Constructed by FewSTAB}\label{sec:more-visualizations}
FewSTAB does not construct tasks based on a specific model. Hence,  FewSTAB is a fair evaluation framework for different FSC methods, and the tasks constructed by FewSTAB can be used to reveal few-shot classifiers' varied degrees of robustness to spurious bias.

\begin{figure*}[htbp]
    \centering
    \includegraphics[width=0.9\linewidth]{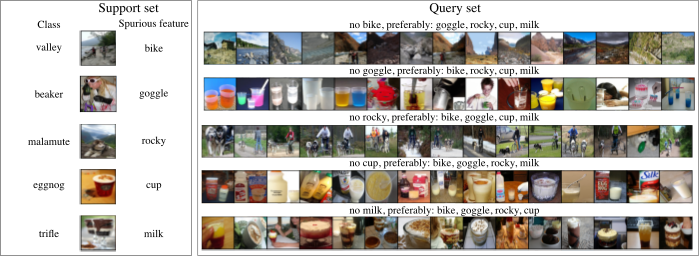}
    \caption{A 5-way 1-shot task constructed by our FewSTAB using samples from the tieredImageNet dataset. Note that due to the limited capacity of a VLM, the attributes may not well align with human understandings.}
    \label{fig:task-example-tiered}
\end{figure*}

\begin{figure*}[htbp]
    \centering
    \includegraphics[width=0.9\linewidth]{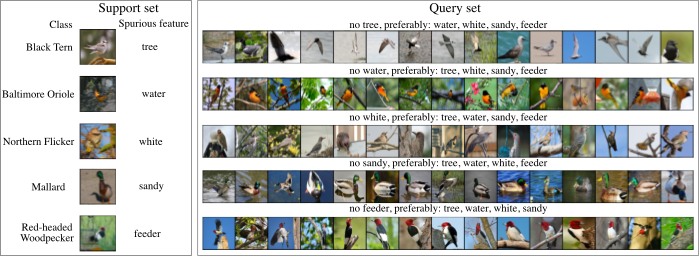}
    \caption{A 5-way 1-shot task constructed by our FewSTAB using samples from the CUB-200 dataset. Note that due to the limited capacity of a VLM, the attributes may not well align with human understandings.}
    \label{fig:task-example-cub}
\end{figure*}

We show a 5-way 1-shot task constructed by FewSTAB using samples from the tieredImageNet and CUB-200 datasets in \cref{fig:task-example-tiered} and \cref{fig:task-example-cub}, respectively. Query samples for each class are constructed such that they do not contain the spurious attribute from the support set sample of the same class but contain spurious attributes from support set samples of other classes. For example, in \cref{fig:task-example-tiered}, the class \texttt{malamute} has a support set sample with a \texttt{rocky} background, but most of its query samples have a \texttt{bike} which is the spurious attribute from the support set sample of the \texttt{valley} class.
Moreover, in \cref{fig:task-example-cub}, the class \texttt{Mallard} has a support set sample with a \texttt{sandy} background, but its query samples all have a \texttt{water} background similar to that in the support set sample of the \texttt{Baltimore Oriole} class. Note that the sample selection may not be ideal due to the limited capacity of VLMs. For example, in \cref{fig:task-example-tiered}, some query images of the class \texttt{eggnog} have the spurious attribute \texttt{cup} which also appears in the support set image of the class, leading to a high accuracy on these query samples for a model that relies on this spurious attribute. However, this does not affect our evaluation of different FSC methods on their robustness to spurious bias since the same set of tasks is used to evaluate different FSC methods. Moreover, our metric, wAcc-A, measures the worst per-class classification accuracy over FSC tasks, making our evaluation robust to the sampling noise caused by a VLM.

\end{document}